\pdfoutput=1

\documentclass[11pt]{article}

\usepackage{EMNLP2022}

\usepackage{times}
\usepackage{latexsym}

\usepackage[T1]{fontenc}

\usepackage[utf8]{inputenc}

\usepackage{microtype}

\usepackage{inconsolata}

\usepackage{graphicx} 
\usepackage{natbib}  
\usepackage{caption} 
\usepackage{algorithm}
\usepackage{algpseudocode}
\usepackage{algorithmicx}
\usepackage{amsfonts}

\newcommand{\texttu}[1]{\texttt{\uppercase{#1}}}

\newcommand{\benchmark}{\textsc{IELM}}

\newcommand{\hide}[1]{}

\usepackage{bibentry}

\usepackage{graphics}
\usepackage{hyperref}
\usepackage{url}
\usepackage{multirow}

\usepackage{colortbl}
\usepackage{makecell}

\usepackage{booktabs}
\usepackage{arydshln}
\usepackage{enumitem}

\definecolor{mediumelectricblue}{rgb}{0.01, 0.31, 0.59}

\newcommand{\comm}[1]{}
\newcommand{\ours}{(\textcolor{mediumelectricblue}{\small ours})}
\newcommand{\oursnormal}{\textcolor{mediumelectricblue}{Ours}}

\newif\ifshowcomment
\showcommentfalse

\ifshowcomment
    
    \newcommand{\dawn}[1]{\textcolor{purple}{[{Dawn: #1}]}}
    \newcommand{\xiao}[1]{\textcolor{red}{[Xiao: #1]}}
    \newcommand{\focus}[1]{\textcolor{red}{#1}}
\else
    
    \newcommand{\dawn}[1]{}
    \newcommand{\xiao}[1]{}
    \newcommand{\focus}[1]{}
\fi

\usepackage{mdwlist}
\usepackage[list=true]{subcaption}
\makeatletter
\newcommand\footnoteref[1]{\protected@xdef\@thefnmark{\ref{#1}}\@footnotemark}
\makeatother

\title{\benchmark: An Open Information Extraction Benchmark for \\Pre-Trained Language Models}
\author {
    Chenguang Wang\textsuperscript{\rm 1}, Xiao Liu\textsuperscript{\rm 2}, Dawn Song\textsuperscript{\rm 3}\\
    \textsuperscript{\rm 1}Washington University in St. Louis, \textsuperscript{\rm 2}Tsinghua University, \textsuperscript{\rm 3}UC Berkeley\\
    \texttt{chenguangwang@wustl.edu}, \texttt{liuxiao21@mails.tsinghua.edu.cn}, \\
    \texttt{dawnsong@berkeley.edu}
}
  
\begin{document}
\maketitle
\begin{abstract}
We introduce a new open information extraction (OIE) benchmark for pre-trained language models (LM). Recent studies have demonstrated that pre-trained LMs, such as BERT and GPT, may store linguistic and relational knowledge. In particular, LMs are able to answer ``fill-in-the-blank'' questions when given a pre-defined relation category. Instead of focusing on pre-defined relations, we create an OIE benchmark aiming to fully examine the open relational information present in the pre-trained LMs. We accomplish this by turning pre-trained LMs into zero-shot OIE systems. Surprisingly, pre-trained LMs are able to obtain competitive performance on both standard OIE datasets (CaRB and Re-OIE2016) and two new large-scale factual OIE datasets (TAC KBP-OIE and Wikidata-OIE) that we establish via distant supervision. For instance, the zero-shot pre-trained LMs outperform the F1 score of the state-of-the-art supervised OIE methods on our factual OIE datasets without needing to use any training sets.\footnote{\label{ft:opensource}Our code and datasets are available at \url{https://github.com/cgraywang/IELM}.}
\end{abstract}

\section{Introduction}

Pre-trained language models (LM), such as BERT~\citep{devlin2018bert} and GPT-3~\citep{brown2020language}, have revolutionized NLP over the last several years and advanced the state-of-the-art results in a wide set of downstream NLP tasks. Recent studies show that a considerable amount of linguistic~\citep{hewitt2019structural,clark2019does} and relational knowledge~\citep{petroni2019language,talmor2019olmpics,jiang2020can,petroni2020context} has been captured by the pre-trained LMs via pre-training on large-scale textual corpora. These approaches often design ``fill-in-the-blank'' questions based on pre-defined relations. For example, a question ``Bob Dylan was born in \_'' is manually created for the LMs to answer the ``birthplace'' relation of ``Bob Dylan''. 

Most existing approaches that evaluate what pre-trained LMs have learned are based on benchmarks with pre-defined relation categories. Yet, the benchmarks present two limitations. First, most benchmarks only cover a limited number of pre-defined relations. Therefore, it is unclear whether the pre-trained LMs have stored general open relation information. For example, the Google-RE in LAMA benchmark~\citep{petroni2019language} includes only three relations (i.e., ``birthplace'', ``birthdate'', and ``deathplace''), while there are hundreds of relations available in the real world scenario. Second, a majority of benchmarks evaluate LMs in a close manner. This means that the gold relation is given to the models. For example, ``was born in'' is given as the model's input. Besides, the existing benchmarks often provide a single gold relation for each input sentence. However, an input sentence may indicate multiple relations, e.g., containing both ``birthplace'' and ``birthdate'' information about an argument or entity. We are curious: instead of the limited relational information, can we systematically benchmark the general information stored in the pre-trained LMs?

\begin{figure}
    \centering
    \includegraphics[width=\linewidth]{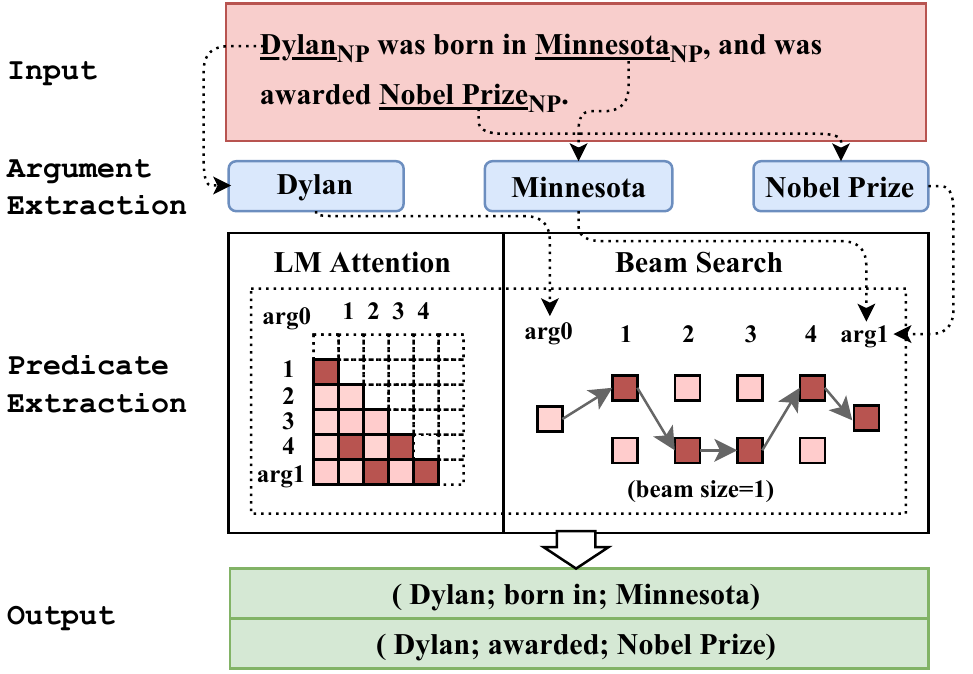}
    \caption{{\small Summary of our approach. The zero-shot open information extraction system takes a noun phrase (NP) chunked sentence as input, and outputs a set of triples. The approach first conducts argument extraction by encoding NPs as argument pairs, then performs predicate extraction via decoding using the parameters (i.e., attention scores) of the pre-trained language models. The output extractions are then evaluated on our \benchmark\ benchmark.}\xiao{update}
      \label{fig:overview}}
\end{figure}

In this work, we set up a new open information extraction (OIE) benchmark, called \benchmark, towards testing the general and open relational information stored in pre-trained LMs. We refer to OIE as it is a task that is designed to extract open relations from massive corpora without requiring a pre-defined relation category. As shown in Figure~\ref{fig:overview}, we successfully convert pre-trained LMs to zero-shot OIE systems. We apply them to two standard OIE datasets, including CaRB~\citep{bhardwaj2019carb} and Re-OIE2016~\citep{stanovsky2016creating,zhan2020span}, as well as two new large-scale factual OIE datasets in our \benchmark\ benchmark. We show that the zero-shot pre-trained LMs outperform the fully supervised state-of-the-arts on factual OIE datasets. Standard OIE datasets rely on human annotations and often consist of thousands of gold triples and sentences. Unlike those datasets, we create two large-scale OIE datasets, namely TAC KBP-OIE and Wikidata-OIE, via distant supervision from knowledge graphs. For example, Wikidata-OIE is constructed via aligning English Wikipedia to Wikidata triples, resulting in millions of triples and documents. The design of zero-shot LMs for OIE is important: by encoding the noun chunks as arguments in the input, we only make use of the parameters of pre-trained LMs to decode the predicates (or relations) between the arguments. To the best of our knowledge, this is the first attempt to systematically evaluate pre-trained LMs in a zero-shot OIE setting. To summarize, our key contributions are the following.
\begin{enumerate}[leftmargin=*]
    \item We benchmark the general relational information in pre-trained LMs on our \benchmark\ benchmark. Besides two standard OIE datasets (CaRB and Re-OIE2016), we also create two large-scale factual OIE datasets for our benchmark. The two new OIE datasets are called TAC KBP-OIE and Wikidata-OIE, which are constructed via distant supervision from two knowledge graphs (TAC KBP and Wikidata). Our benchmark is a general OIE benchmark, helping the development of future OIE systems.
    \item We enable the zero-shot capabilities of pre-trained LMs for OIE by encoding the arguments in the input and decoding predicates using the parameters of pre-trained LMs. The pre-trained LMs are particularly good at recovering factual arguments and predicates.
    \item We test the OIE performance of 6 pre-trained LMs (BERT and GPT-2~\citep{radford2019language} families) and 14 OIE systems on \benchmark\ benchmark. The zero-shot LMs achieve state-of-the-art OIE performance on TAC KBP-OIE and Wikidata-OIE, even outperforming fully supervised OIE systems.
\end{enumerate}

\section{Language Models as Zero-Shot Information Extractors}
\label{sec:app}

For open information extraction (OIE), we take an input as a NP-chunked sentence and output a set of triples. Below is an example.
\begin{enumerate*}
    \item[] {\bf Input} \underline{Dylan}$_{\rm NP}$ was born in \underline{Minnesota}$_{\rm NP}$, and was awarded \underline{Nobel Prize}$_{\rm NP}$.
    \item[] {\bf Output} (Dylan; born in; Minnesota), (Dylan; awarded; Nobel Prize).
\end{enumerate*}
${\rm NP}$ denotes the noun phrase.

\subsection{Argument Extraction} 
Follow traditional linguistic OIE systems such as Stanford OpenIE~\citep{angeli2015leveraging} and OpenIE5~\citep{saha2017bootstrapping,saha2018open}, we treat each NP pair as an argument pair (e.g., ``Dylan'' and ``Minnesota''). We then utilize the parameters of LMs to extract the predicates (e.g., ``born in'') between the pair in the input as below.

\begin{figure*}
\centering
\subcaptionbox{{\small Predicate extraction example.}}{\includegraphics[width=0.45\textwidth]{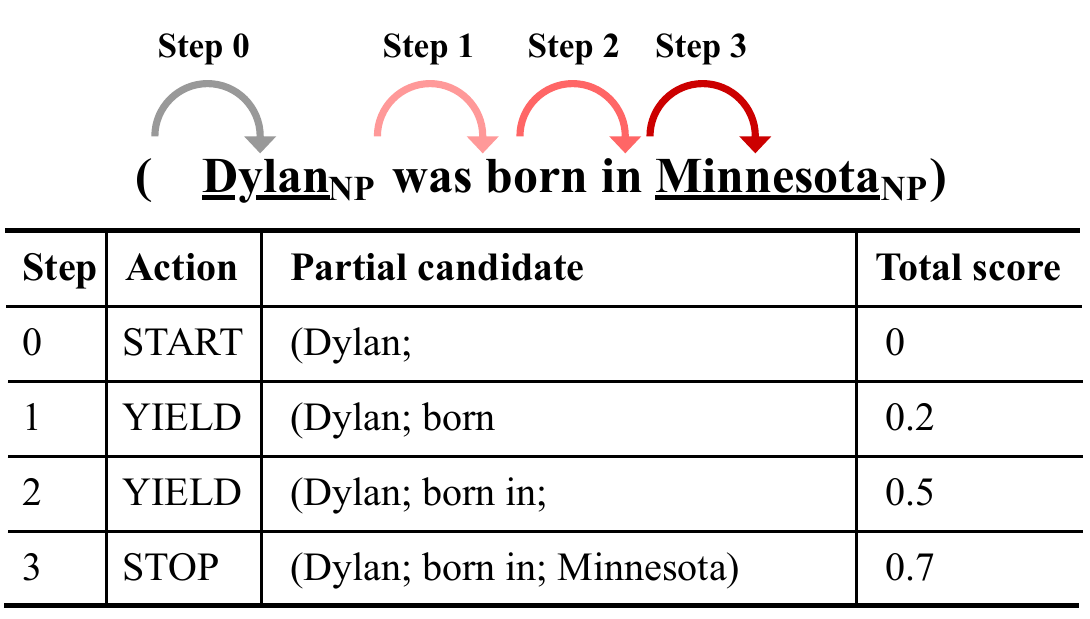}}%
\hspace{0.2in}
\subcaptionbox{{\small Attention matrix.}}{\includegraphics[width=0.27\textwidth]{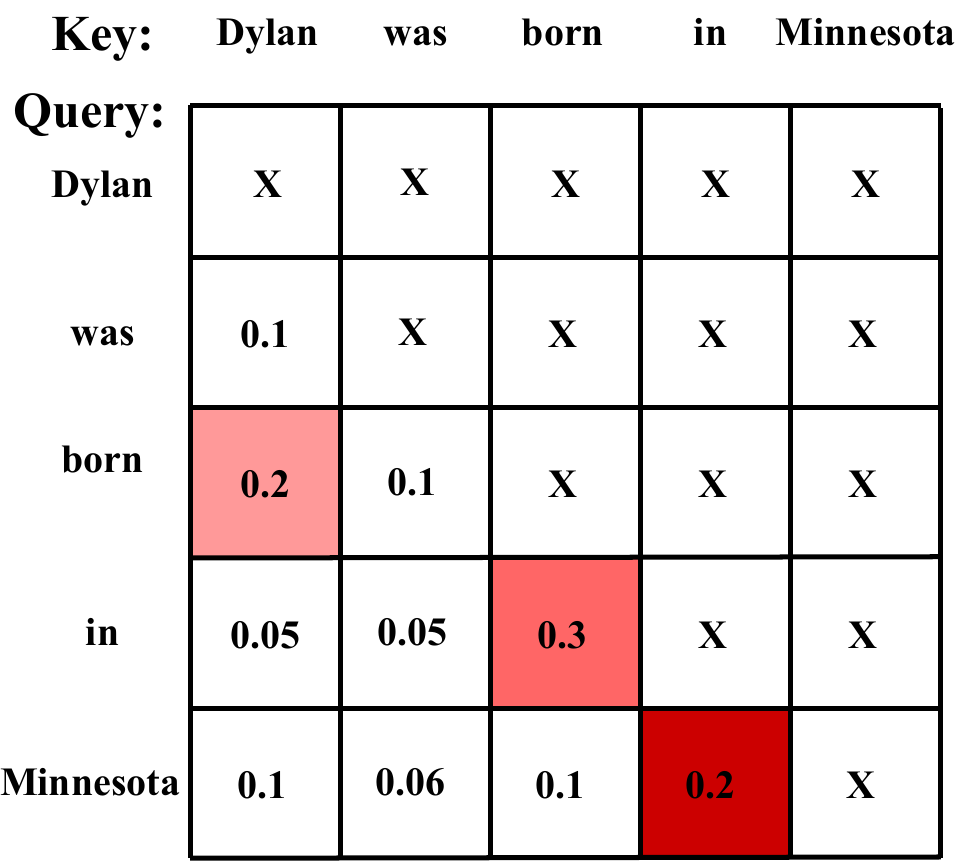}}%
\caption{{\small Illustration of predicate extraction with a pre-trained language model (LM). The upper part of (a) represents the general search steps of producing the triple {\sl (Dylan; born in; Minnesota)} from the input ``\underline{Dylan}$_{\rm NP}$ was born in \underline{Minnesota}$_{\rm NP}$'' encoded with argument noun phrases (NP). The lower portion shows the corresponding step-by-step process. (b) shows the attention scores generated through the forward pass of the LM over the corresponding input.}\xiao{update}}
\label{fig:search}
\end{figure*}


\subsection{Predicate Extraction}
\label{sec:predicate}
The predicate extraction problem is formulated as extracting a set of sequences in the input that are associated with an argument pair. We particularly use the attention scores in a pre-trained LM to measure the relevance between a sequence and the argument pair. We frame the process as a search problem. Given an argument pair, we aim to search for the sequences with the largest attention scores connecting the pair. To compute a score for every possible sequence is computationally expensive especially when the sequence length is large, the exhaustive search is therefore intractable. We adapt beam search as an approximate strategy to efficiently explore the search space. Beam search maintains the $k$-best candidates. This means the time cost of beam search does not depend on the sequence length, but on the size of the beam and the average length of the candidates. In general, the beam search starts with the first argument (e.g., ``Dylan''). At each step, beam search simply selects top-$k$ next tokens with the largest attention scores, and just keeps $k$ partial candidates with the highest scores, where $k$ is the beam size. When a candidate produces the second argument (e.g., ``Minnesota''), the candidate is complete.

We show a running example as follows. Let's first consider the search from left to right with beam size equal to 1. An example search process is shown in Figure~\ref{fig:search}. Given an argument pair ``Dylan'' and ``Minnesota'', at each step, the search performs one of the following actions:

\begin{itemize}[leftmargin=*]
    \item {\texttu{start}} the search from first argument. The first argument is added as an initial candidate into the beam. In Figure~\ref{fig:search}(a), at step 0, ``Dylan'' is added into the beam. The total attention score is initialized to 0.
    \item {\texttu{yield}} a new partial candidate in the beam if the current candidate has not reached the second argument. This action conducts the following: The next largest attended token is appended to the end of the current candidate to yield the new candidate. The total score is increased by the associated attention score. At step 1 of Figure~\ref{fig:search}(a), ``born'' is appended to the current candidate to yield the partial candidate, since ``born'' has the largest attention score (0.2 as highlighted in Figure~\ref{fig:search}(b)) with ``Dylan'' in the attention matrix. The total score becomes 0.2. Note that we only consider the single head attention in this example for simplicity. ``x'' in Figure~\ref{fig:search}(b) marks the tokens (prior to the current token) that are not considered in the search to prevent searching backward. Step 2 takes the same action, and the score becomes 0.5.
    \item {\texttu{stop}} the search step if the candidate has reached the second argument, then add the candidate as a valid triple into the beam. As beam size equals to 1, {\sl (Dylan; born in; Minnesota)} is returned for the given pair. The final score of the triple is 0.7. 
\end{itemize}

We also notice triples are often in reverse order in the sentence, thus enabling bidirectionality by running the algorithm in both directions (left to right and right to left). We merge the subwords as words, and only consider word-level attention. The beam search is implemented by the breadth-first search, which is efficient as the time complexity is $O(k \cdot d)$. $d$ is the maximum depth of the search tree.

\section{The \benchmark\ Benchmark}

\subsection{Datasets}
\label{sec:datasets}

\subsubsection{Standard OIE}
We adopt two standard OIE datasets below. 
\paragraph{CaRB}
CaRB~\citep{bhardwaj2019carb} is a crowdsourced OIE dataset, where the input sentences are from the OIE2016~\citep{stanovsky2016creating}.

\paragraph{Re-OIE2016}
Re-OIE2016~\citep{zhan2020span} is also generated based on the input sentences in the OIE2016, and is further enhanced by human annotations.

\subsubsection{Factual OIE}
In addition, we introduce two large-scale factual OIE datasets based on knowledge graphs (KG).

\paragraph{TAC KBP-OIE}
TAC Knowledge Base Population (KBP) Slot Filling is a task to search a document collection to fill in a target entity for predefined relations (slots) with a given entity in a reference KG. We adapt the dataset as an OIE dataset. In particular, we use a document sub-collection of the TAC KBP 2013 task~\citep{surdeanu2013overview} as the input, and use the official human annotations regarding the documents as gold extractions.

\paragraph{Wikidata-OIE}
Besides TAC KBP-OIE, we create a larger factual OIE dataset based on the English Wikipedia. Different from TAC KBP, there are no gold triple annotations for Wikipedia. Since a large amount of Wikidata triples are from English Wikipedia, we create the dataset using distant supervision~\cite{zhang2017position} by aligning Wikidata triples to Wikipedia text. We employ an unsupervised entity linker based on a pre-built mention-to-entity dictionary~\citep{spitkovsky2012cross} to extract potential gold arguments for scalability considerations. The entity linker links an arbitrary entity mention in a sentence to a Wikipedia anchor, which is further linked to a Wikidata entity. For each sentence in Wikipedia articles containing two linked arguments, if there is a Wikidata triple describing a relation holding the two arguments, we denote the Wikidata triple as a gold extraction.

Unlike TAC KBP-OIE which is built based on human annotations, Wikidata-OIE is derived from automatic annotations. Therefore, we evaluate our unsupervised entity linker on the standard AIDA benchmark~\cite{hoffart2011robust} consisting of Wikipedia entities. Table~\ref{tab:el} shows that it significantly improves the unsupervised performance~\citep{spitkovsky2012cross} and reaches competitiveness with a supervised method~\cite{kolitsas2018end}. Given the scale of Wikidata-OIE, we sacrifice acceptable effectiveness for efficiency.

\begin{table}[]
\centering
\resizebox{0.7\linewidth}{!}{
\begin{tabular}{@{}lcc@{}}
\toprule
\textbf{Method} & \multicolumn{2}{c}{\textbf{AIDA}} \\
                & dev             & test            \\ \midrule
\citealt{spitkovsky2012cross}                & 26.0            & 28.2            \\
\citealt{kolitsas2018end}$^*$                & -               & 82.4            \\
\oursnormal           & 63.8            & 64.5            \\ \bottomrule
\end{tabular}
}
\caption{{Evaluation of unsupervised entity linking of Wikidata-OIE on AIDA benchmark. An asterisk (*) indicates a supervised method.}}
\label{tab:el}
\end{table}

The statistics of the datasets are shown in Table~\ref{tab:data}. For CaRB and Re-OIE2016, we report the statistics of the corresponding test sets. We include a dataset comparison in Appendix~\ref{sec:appendixdatasetcomp}.

\begin{table}[t]
\centering
\resizebox{\linewidth}{!}
  {
  \begin{tabular} {l | c | c | c | c}
        \toprule
    {\bf Dataset} & {\bf \# of triples} & {\bf \# of arguments} & {\bf \# of predicates} &  {\bf \# of documents} \\
    \hline 
        Re-OIE2016                      &   1,508      & 3,328  & 1,506           & 595 \\
        CaRB                      &  2,715       & 6,226  & 2,715           & 641 \\
        TAC KBP-OIE                              & 27,655     & 39,661 &    41                   & 3,877,207 \\
        Wikidata-OIE                      & 27,368,562       & 6,047,494 & 1,156           & 6,047,494 \\
    \bottomrule
  \end{tabular}
  }
\caption{{Dataset statistics of the \benchmark\ benchmark.\xiao{*Two input sentences of the Re-OIE2016 have no extraction groundtruth.}}}
\label{tab:data}
\end{table}


\subsection{Pre-Trained Language Models for OIE}
\paragraph{Unidirectional Language Models}
Given an input sequence $\mathbf{x} = \{x_1, x_2, \dots, x_N\}$, unidirectional LMs assign a joint probability to the sequence by factorizing it as $p(\mathbf{x}) = \prod_t p(x_t | x_{t-1},\dots,x_1)$, where $p(x_t | x_{t-1},\dots,x_1) = \sigma(\mathbf{W}\mathbf{h}_t+\mathbf{b})$. $\mathbf{h}_t$ is the output vector of a neural network at position $t$.

We consider GPT-2~\citep{radford2019language}, where $\mathbf{h}_t$ is produced by Transformer decoders~\citep{vaswani2017attention}. GPT-2 is pre-trained on WebText containing 40GB of text. We explore all four pre-trained GPT-2s with different model sizes: {GPT-2} (117M), {GPT-2$_{\rm MEDIUM}$} (345M), {GPT-2$_{\rm LARGE}$} (774M), and {GPT-2$_{\rm XL}$} (1558M).

\paragraph{Bidirectional Language Models}
Different from unidirectional LMs that predict the next word given the previous words, bidirectional LMs take both left and right context of the target word into consideration, formally, $p(x_t) = p(x_t | x_1, \dots, x_{t-1}, x_{t+1}, \dots, x_N)$.

We use BERT~\citep{devlin2018bert} that enables bidirectional context modeling via a masked LM objective and utilizing the Transformer architecture. BERT is pre-trained on BooksCorpus and English Wikipedia. We use both its pre-trained settings: BERT$_{\rm BASE}$ (109M) and BERT$_{\rm LARGE}$ (335M).

\subsection{Comparison Methods}

We compare our method with a wide set of OIE systems including both neural and traditional linguistic OIE systems.
Most OIE systems are based on supervised learning, which are indicated with asterisks (*) in Table~\ref{tab:resultall}. We provide details of the comparison systems in Appendix~\ref{sec:appendixcompsys}.

\subsection{Evaluation Method}
\label{sec:evalmethod}
\subsubsection{Standard OIE}
On CaRB and Re-OIE2016, we follow the original evaluation proposed in \citep{bhardwaj2019carb} and \citep{stanovsky2016creating,zhan2020span}, and report precision, recall, F1, area under the curve (AUC) for compared OIE systems. AUC is calculated from a plot of the precision and recall values for all potential confidence thresholds. The F1 is the maximum value among the precision-recall pairs. We follow the matching function proposed for each dataset, i.e., lexical match for Re-OIE2016, and tuple match for CaRB. The CaRB evaluation function is stricter as it penalizes long extractions.

\subsubsection{Factual OIE}
We report precision, recall, and F1 of the OIE systems on two large-scale factual OIE datasets: TAC KBP-OIE and Wikidata-OIE. We introduce {\sl exact match} as the matching function for them as below.

\paragraph{Matching Function} The matching functions for standard OIE datasets are generally flexible. For example, the lexical match of Re-OIE2016 judges an argument or predicate as correct if and only if it includes the syntactic head of the gold argument or predicate. Unlike these matching functions, our exact matching function requires both arguments and predicates are linked to the gold extractions. 

For TAC KBP-OIE, we judge an argument to be correct if and only if it matches the name of the gold argument and the span position of the gold argument in the sentence. The main challenge is how to properly link a predicate, since there are often many ways to express it. We follow Stanford OpenIE~\citep{angeli2015leveraging} to produce the predicate mapping between the OIE relations and TAC KBP predicates. A predicate is correct if the pair of OIE relation and gold predicate exists in the predicate mapping. The predicate mapping is constructed in two steps. First, a collection of predicate mappings was constructed by a single annotator in approximately a day. Second, predicate mappings were finalized through the following learning procedure. This process matches OIE relations to the TAC KBP predicates by searching for co-occurrent relations in a large distantly-labeled corpus, and decides pairs of OIE relations and TAC KBP predicates that have a high PMI$^2$. The basic idea is that the more often the argument pairs of the triples and TAC KBP triples are linked, the more likely the corresponding relations or predicates are linked to each other. Example predicate mappings are shown in Appendix~\ref{sec:appendixmapping}.

For Wikidata-OIE, we link an argument based on the entity linker used in Wikidata-OIE construction (Sec.~\ref{sec:datasets}). 
An argument is correct if the linked argument matches the name of the gold argument and the span position of the gold argument in the sentence. The predicate mapping is bootstrapped from TAC KBP-OIE's mapping. In addition, we normalize each predicate phrase of the triples by lemmatization, and removing inflection, auxiliary verbs, adjectives, adverbs. One author manually filters out the bad predicate pairs. This process takes approximately a day. A predicate is correct if the OIE to gold predicate pair exists in the bootstrapped predicate mapping. An annotator randomly subsampled and checked 100 aligned triple-sentence pairs and concluded with a 93\% accuracy of extracted triples.

\paragraph{Metrics} We use the official scorer of TAC KBP Slot Filling 2013 to calculate precision, recall, and F1 for TAC KBP-OIE. Besides, like previous OIE systems, for LMs, we adopt two constraints from ReVerb~\citep{fader2011identifying}, the antecessor of OpenIE5: (\expandafter{\romannumeral1}) the frequency of predicates must be above a threshold aiming to avoid triples to be over-specified, and (\expandafter{\romannumeral2}) a predicate must be a contiguous sequence in the sentence avoiding predicates that have no meaningful interpretation. We set these parameters according to Sec.~\ref{sec:para}.

We only report precision, recall, and F1 based on the parameter study in Sec.~\ref{sec:para} for pre-trained LMs on the \benchmark\ benchmark. We do not compute AUC as pre-trained LMs are treated as zero-shot OIE systems. We therefore do not tune the results with respect to different confidence. Our main focus is to benchmark the OIE performance of the LMs under a unified setting. Another reason is that it is computationally expensive to get the AUC on the two large-scale datasets: TAC KBP-OIE and Wikidata-OIE. We also do not report AUC for the compared OIE systems on TAC KBP-OIE and Wikidata-OIE. Instead, we use the confidence threshold that obtains the best F1 on Re-OIE2016 to compute the scores.

\begin{table*}[]
\resizebox{\textwidth}{!}{%
\begin{tabular}{ll|cccc|cccc|ccc|ccc}
\toprule
\multirow{2}{*}{{\bf Method}} & \multirow{2}{*}{{\bf \#Params}} & \multicolumn{4}{c|}{{\bf CaRB}}  & \multicolumn{4}{c|}{{\bf Re-OIE2016}} & \multicolumn{3}{c|}{\bf{TAC KBP-OIE}} & \multicolumn{3}{c}{\bf{Wikidata-OIE}}   \\
                        &                           & P    & R    & F1   & AUC  & P      & R     & F1    & AUC   & P     & R     & F1     & P      & R      & F1     \\
\midrule
MinIE~\citep{gashteovski2017minie}                   & -                         & -    & -    & 41.9 & -    & 48.2      & 71.1     & 57.5     & 47.2     &  -     &  -     & -      &  -      &  -      & -          \\
ClausIE~\citep{del2013clausie}                 & -                         & -    & -    & 44.9 & 22.4 & -      & -     & 64.2  & 46.4  &   -    &   -    &  -     &  -   &   -     &  -      \\
OLLIE~\citep{schmitz2012open}                   & -                         & -    & -    & 41.1 & 22.4 & -      & -     & 49.5  & 31.3  &  -     &   -    &  -     &   -  &   -     &  -      \\
PropS~\citep{stanovsky2016getting}                   & -                         & -    & -    & 31.9 & 12.6 & -      & -     & 64.2  & 43.3  &  -     & -      &  -     & -    &   -     &  -     \\
OpenIE4~\citep{christensen2011analysis}                & -                         & -    & -    & 48.8 & 27.2 & -      & -     & 68.3  & 50.9  & 52.9     & 14.2     & 22.4        & 30.6  & 14.0  & 19.2    \\
OpenIE5~\citep{saha2017bootstrapping,saha2018open}                & -                         & -    & -    & 48.0 & 25.0 & 61.1      & 76.5     & 67.9     & 45.8     & 57.0 & 14.6 & 23.2   & 26.3 & 14.4 & 18.6   \\
Stanford OpenIE$^*$~\citep{angeli2015leveraging}         & -                         & -    & -    & 23.0 & 13.4 & -      & -     & 16.7  & 11.5  & 61.6 & 17.4 & 27.1 & 23.3  & 13.1  & 16.8    \\
 \hdashline
SenseOIE$^*$~\citep{roy2019supervising}                & 640K                         & -    & -    & 28.2 & -    & -      & -     & -     & -     &  -     &    -   &   -    &  -   &   -     &     -    \\
SpanOIE$^*$~\citep{zhan2020span}                 & 963K                         &  60.9 & 41.6 & 49.4 & 30.0 & 79.7   & 74.5  & 77.0  & 65.8  &   -    &     -  &     -  &  -   &     -   &      -   \\
NeuralOIE$^*$~\citep{cui2018neural}               & 5M                         & -    & -    & 51.6 & 32.8 & 79.2      & 77.5     & 78.4     & 73.0     &   -    &     -  &     -  &   -  &    -    &    -    \\
Multi$^2$OIE$^*$~\citep{ro2020multi}               &       110M                  &  60.9 &  45.8 & 52.3 & 32.6 &  86.9   &  81.0  &  83.9  &  74.6  &     -  & -      &    -   &    - &    -    &    -    \\
RnnOIE$^*$~\citep{stanovsky2018supervised}                  & 965K                         & 55.6 & 40.2 & 46.7 & 26.8 & 84.2   & 73.9  &  78.7  & 68.3  &  50.0    &  14.6  & 22.6  & 29.9  & 15.9  & 20.7    \\
IMOJIE$^*$~\citep{kolluru2020imojie}                  & 110M                         &  64.7    &  45.6    &  53.5 &  33.3 &  88.1   & 67.1  & 76.2  & 63.1  &    58.2   &   14.9    &    23.8     &     31.2   &   16.2     &    21.3     \\
OpenIE6$^*$~\citep{kolluru2020openie6}                & 220M                         & -    & -    &  54.0 &  35.7 & 75.3      &  78.2     & 76.7     &  73.8     & 60.0 & 14.9 & 23.9      & 29.8 & 15.3 & 20.3   \\
 \hdashline
BERT$_{\rm BASE}$ (zero-shot) \ours            & 109M                      & 21.2 & 18.3 & 19.7 & - & 25.9 & 34.0 & 29.4 & - & 61.6 & 18.8 & 28.8    & 32.0 &  18.9 & 23.7   \\
BERT$_{\rm LARGE}$ (zero-shot) \ours           & 335M                      & 22.4 & 20.2 & 21.2 & - & 30.7 & 38.5 & 34.1 & - & 61.7 & 19.0 & 29.1   & 32.3  &  19.1  &  24.0    \\
GPT-2 (zero-shot)  \ours                & 117M                      & 23.1 & 19.8 & 21.3 & - & 25.1 & 38.3 & 30.3 & - & 61.6 & 18.2 & 28.1   & 32.4      & 18.1      & 23.2       \\
GPT-2$_{\rm MEDIUM}$ (zero-shot) \ours         & 345M                      & 23.7 & 20.0 & 21.7 & - & 26.8 & 39.9 & 32.0 & - & 62.1 & 18.7 & 28.7    & 33.1      & 18.2      & 23.5       \\
GPT-2$_{\rm LARGE}$ (zero-shot) \ours          & 774M                      & 24.2 & 20.5 & 22.2 & - & 27.4 & 41.6 & 33.0 & - &  62.4 &  19.0 &  29.1   &  34.2      & 18.3      & 23.8        \\
GPT-2$_{\rm XL}$ (zero-shot) \ours             & 1558M                     & 24.5 & 21.0 & 22.7 & - & 29.3 & 43.4 & 35.0 & - &  62.7 &   19.5 &  29.7   &  35.7      & 18.5      &  24.4      \\
\bottomrule
\end{tabular}%
}
\caption{Compare the quality of different OIE systems. An asterisk (*) indicates a supervised method. \xiao{update ?}
}
\label{tab:resultall}
\end{table*}

\section{Results}
\label{sec:exp}

In this section, we show that pre-trained LMs are effective zero-shot OIE systems, and exceed the previous state-of-the-art OIE systems on our large-scale factual OIE datasets in \benchmark\ benchmark. To keep our evaluation as simple as possible, the hyperparameters and settings are shared across datasets. More experimental details are described in the appendix.

\subsection{OIE Results}
\label{sec:oieres}

Table~\ref{tab:resultall} shows the results. While zero-shot OIE systems synthesized by pre-trained LMs obtain notably lower scores compared to previous OIE systems on standard OIE datasets, they outperform the previous OIE systems on factual OIE datasets. We also find that larger LMs obtain improved results on all datasets. For example, BERT$_{\rm LARGE}$ outperforms BERT$_{\rm BASE}$ due to its larger model size. GPT-2s share similar trends. This is because larger LMs store richer relational information. This finding is consistent with previous studies~\citep{petroni2019language,petroni2020context}.

\subsubsection{Standard OIE}

The main reasons for the degraded performance of pre-trained LMs on standard OIE datasets are three-fold. First, the comparison methods mainly involve supervised systems that are trained on OIE datasets, which are denoted with asterisks (*) in Table~\ref{tab:resultall}. Besides the supervised systems, the remaining comparison systems all require human involvement, such as providing linguistic patterns for the extraction. In contrast, the pre-trained LMs are used as zero-shot OIE systems without using any training sets. Second, the zero-shot OIE modules still have room to improve. For example, approximately 30.0\% of the argument extraction errors are due to the spaCy noun chunker. 16.9\% of the gold extractions contain predicates outside the argument pairs. The current predicate extraction only allows searching between the arguments, and thus cannot handle such cases. Third, standard OIE benchmarks such as CaRB and Re-OIE2016 mainly examine the general information extraction capability. The zero-shot approach is not able to recall the information of interest in LMs. We might need a specific module (e.g., ranking) to locate such information. Interestingly, pre-trained LMs achieve comparable performance with supervised Stanford OpenIE. The result indicates pre-trained LMs contain informative patterns that are useful for OIE.

\subsubsection{Factual OIE}

\begin{table*}[t]
\centering
\resizebox{\linewidth}{!}{%
\begin{tabular}{@{}lcccccccccccccccccccc@{}}
\toprule
\textbf{Method}         & MinIE & ClausIE & OLLIE & PropS & OpenIE 4 & OpenIE 5 & \multicolumn{2}{c}{Stanford OpenIE} & SpanOIE & RnnOIE \\ \midrule
\textbf{Sentences/sec.} & 8.9   & 4.0     & 14.5  & 4.6   & 20.1     & 3.1      & \multicolumn{2}{c}{2.5}             & 19.4  & 149.2  \\ \midrule
\textbf{Method}         & NeuralOIE & IMOJIE & Multi$^2$OIE & OpenIE 6 & BERT$_{\rm BASE}$   & BERT$_{\rm LARGE}$    & GPT-2     & GPT-2$_{\rm MEDIUM}$    & GPT-2$_{\rm LARGE}$    & GPT-2$_{\rm XL}$  \\ \midrule
\textbf{Sentences/sec.} & 11.5      & 2.6    & 21.2      & 142.0    & 16.2 & 11.9  & 13.9  & 12.7  & 11.5  & 11.0 \\ \bottomrule
\end{tabular}%
}
\caption{Runtime on Re-OIE2016. \xiao{update ?}}
\label{tab:runtime}
\end{table*}

\comm{
\begin{table}[]
\centering
\resizebox{0.7\linewidth}{!}{%
\begin{tabular}{ll}
\toprule
{\bf Method}          & {\bf Sentences/sec.} \\
\midrule
MinIE           & 8.9            \\
ClausIE         & 4.0            \\
OLLIE           & ?              \\
PropS           & 4.6            \\
OpenIE 4        & 20.1           \\
OpenIE 5        & 3.1            \\
Stanford OpenIE & ?              \\
SenseOIE        & -              \\
SpanOIE         & 19.4           \\
RnnOIE          & 149.2          \\
NeuralOIE       & 11.5           \\
IMOJIE          & 2.6            \\
Multi2OIE       & 21.2              \\
OpenIE 6        & 142.0          \\
BERT\_base      & ?              \\
BERT\_large     & ?              \\
GPT-2           & ?              \\
GPT-2\_medium   & ?              \\
GPT-2\_large    & ?              \\
GPT-2\_xl       & ?              \\
\bottomrule
\end{tabular}%
}
\caption{Runtime on 3,200 sentences from Re-OIE2016.\xiao{update ?}}
\label{tab:runtime}
\end{table}}

As shown in Table~\ref{tab:resultall}, the best zero-shot OIE system based on GPT-2$_{\rm XL}$ obtains a +2.6\% and a +3.1\% absolute F1 improvement on TAC KBP-OIE and Wikidata-OIE respectively over the previous supervised state-of-the-art. Due to the computation cost of OIE systems (Sec.~\ref{sec:runtime}), we only select several best performed OIE systems on the standard OIE datasets for the large-scale OIE experiments including: linguistic OIE systems (OpenIE4, OpenIE5, Stanford OpenIE) and neural OIE systems (RnnOIE, IMOJIE, OpenIE6). 

Compared to the results on standard OIE datasets, pre-trained LMs consistently achieve state-of-the-art performance on both datasets. Both datasets emphasize measuring factual arguments and predicates in the reference KGs. Previous studies~\citep{petroni2019language,petroni2020context} show that LMs have stored a considerable amount of factual information via pre-training on large-scale text. We draw the same conclusion. To the best of our knowledge, our \benchmark\ benchmark is the first benchmark that includes factual OIE datasets. More importantly, both linguistic and neural OIE systems are derived from manually designed linguistic patterns or learned patterns. The result shows that the pre-trained attention weights capture a more flexible set of factual patterns. The result also suggests that our approach is capable of using such patterns.
In order to scale our approach to large-scale datasets, the argument and predicate extraction are both efficient by design. In particular, the beam search for predicate extraction is efficient in exploring the relational sequences in the input sentence. Besides, the attention scores used in the beam search are produced via a single forward pass of the pre-trained LM over the input sentence without fine-tuning.

Moreover, we find that BERT LMs outperform GPT-2 LMs under similar model sizes. On both datasets, BERT$_{\rm BASE}$ performs better than GPT-2 in F1, and BERT$_{\rm LARGE}$ outperforms GPT-2$_{\rm MEDIUM}$ in F1. This is mainly because the recall of BERT LMs is higher than that of corresponding GPT-2 LMs. The result indicates that the Cloze-style loss function (i.e., masked LM) of BERT is more effective and flexible in recovering information than the autoregressive LM objective. We also notice that the precision of GPT-2 LMs is higher than that of BERT LMs. The reason is that the autoregressive LM objective captures more accurate information than Cloze-style loss does by preventing extra noise (e.g., masked tokens) in pre-training.

Pre-trained LMs achieve competitive precision, e.g., the precision is greater than 60\% on TAC KBP-OIE. However, only moderate recalls are obtained. Therefore, improving recall is clearly the future direction. We find that both argument and predicate extraction can be further improved. For example, the main cause of the moderate recall is the incorrect arguments caused by spaCy noun chunks as summarized in Sec.~\ref{sec:error}. Besides, we can incorporate predicates that are not between the argument pairs into the extractions, as we observe a number of gold triples are in inverted sentences. 
We also notice that the F1 gain over previous state-of-the-arts on TAC KBP-OIE is smaller compared to that on Wikidata-OIE. A larger text corpus, e.g., Wikipedia, provides more information. We could improve the recall by running on larger corpora such as WebText2 and Common Crawl~\citep{abs-1910-10683,brown2020language} to collect more triples.

\begin{figure*}
\centering
\subcaptionbox{{\small Beam size.}}{\includegraphics[width=0.28\textwidth]{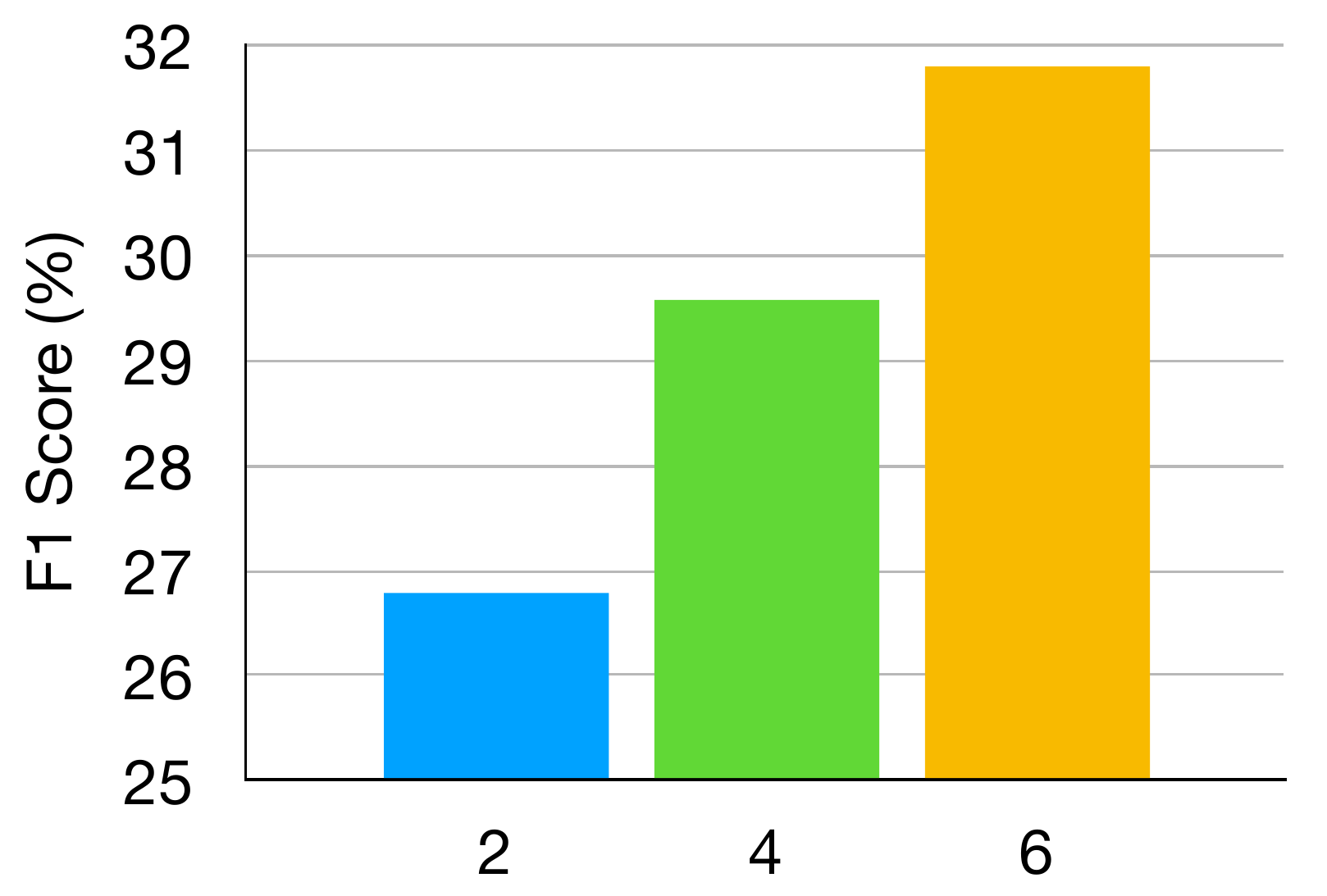}}%
\hspace{0.1in}
\subcaptionbox{{\small Total score threshold.}}{\includegraphics[width=0.28\textwidth]{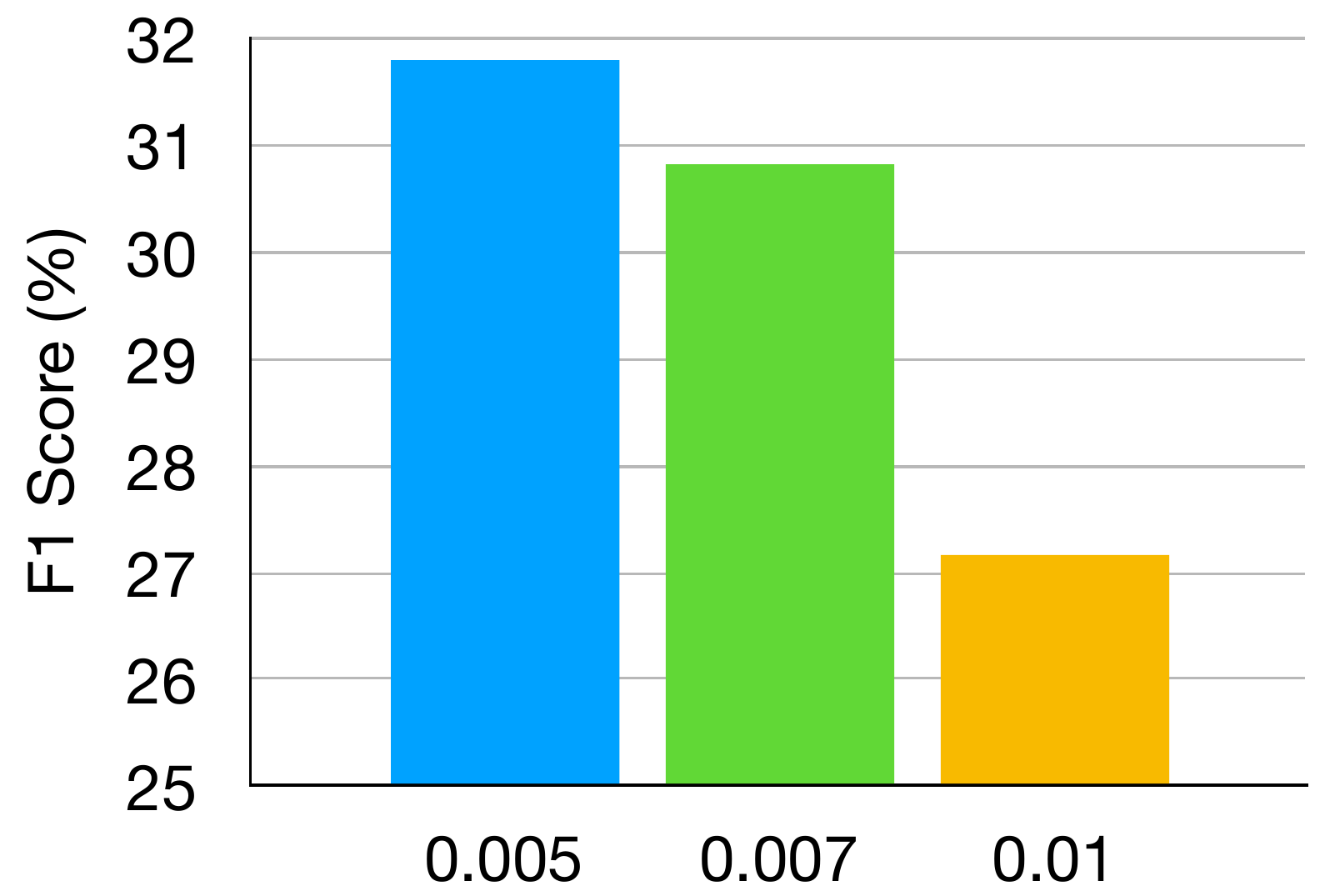}}%
\hspace{0.1in}
\subcaptionbox{{\small Predicate frequency.}}{\includegraphics[width=0.28\textwidth]{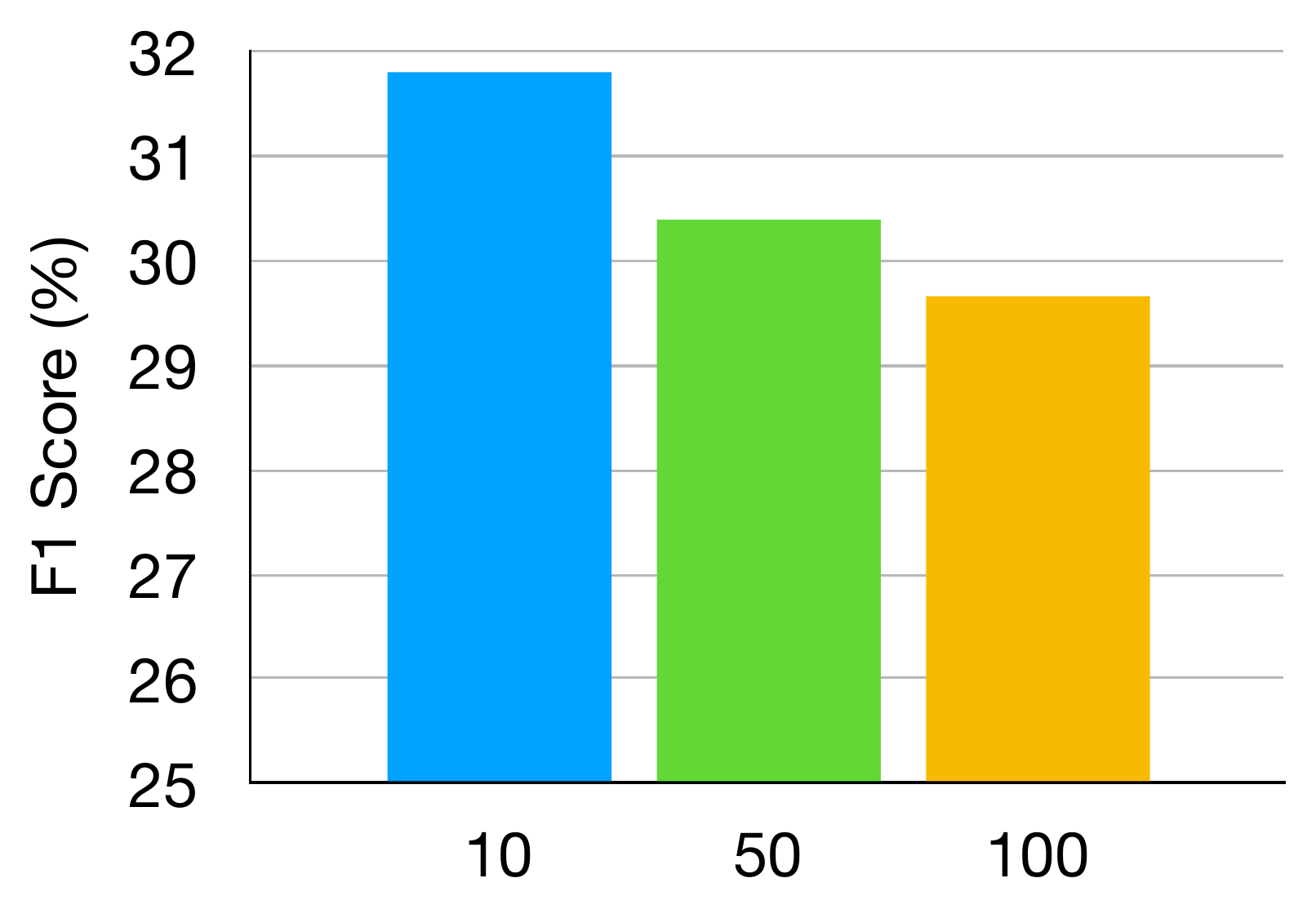}}%
\\
\subcaptionbox{{\small Attention weights layers.}}{\includegraphics[width=0.28\textwidth]{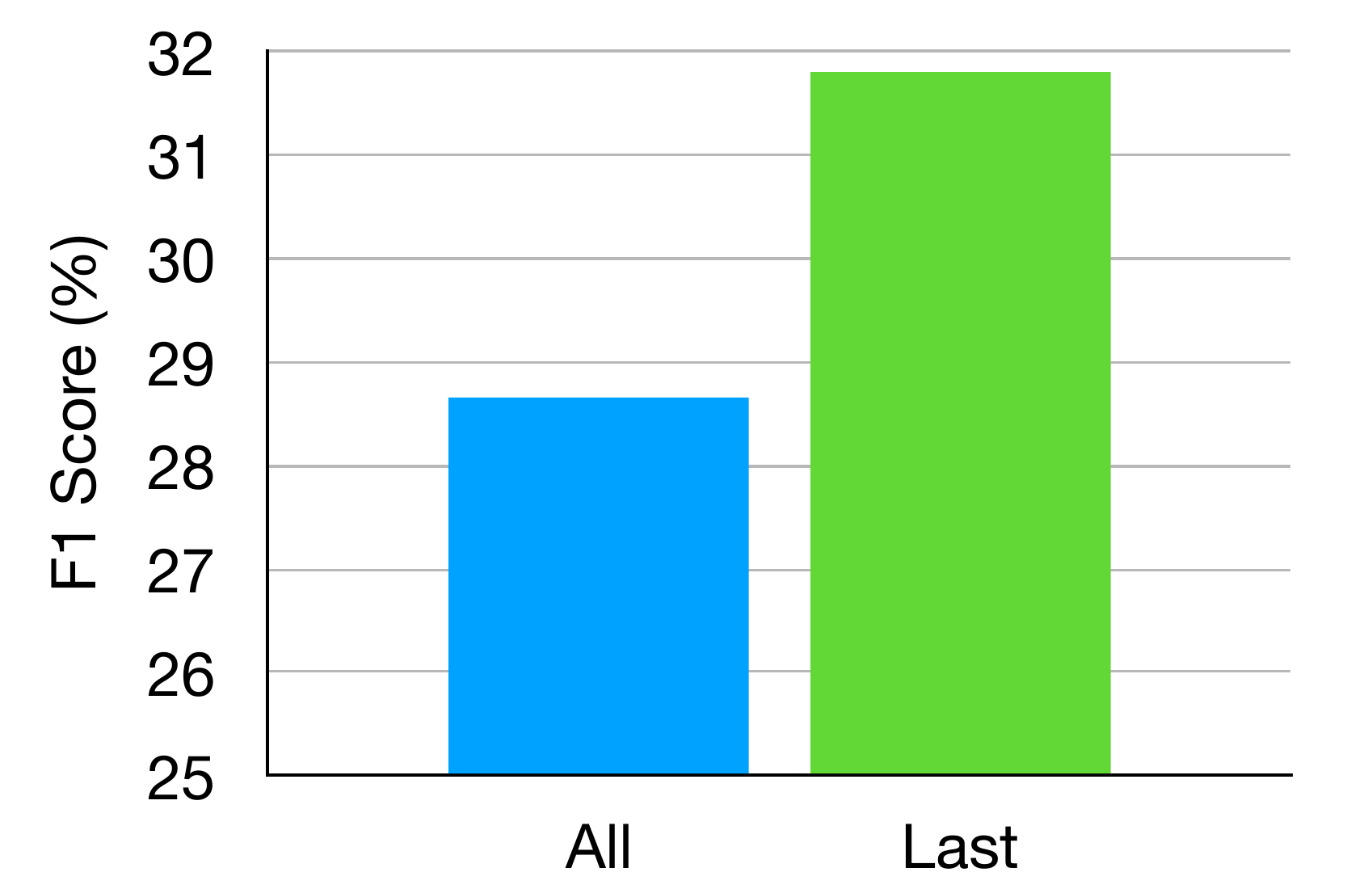}}%
\hspace{0.1in}
\subcaptionbox{{\small Attention weights reduction.}}{\includegraphics[width=0.28\textwidth]{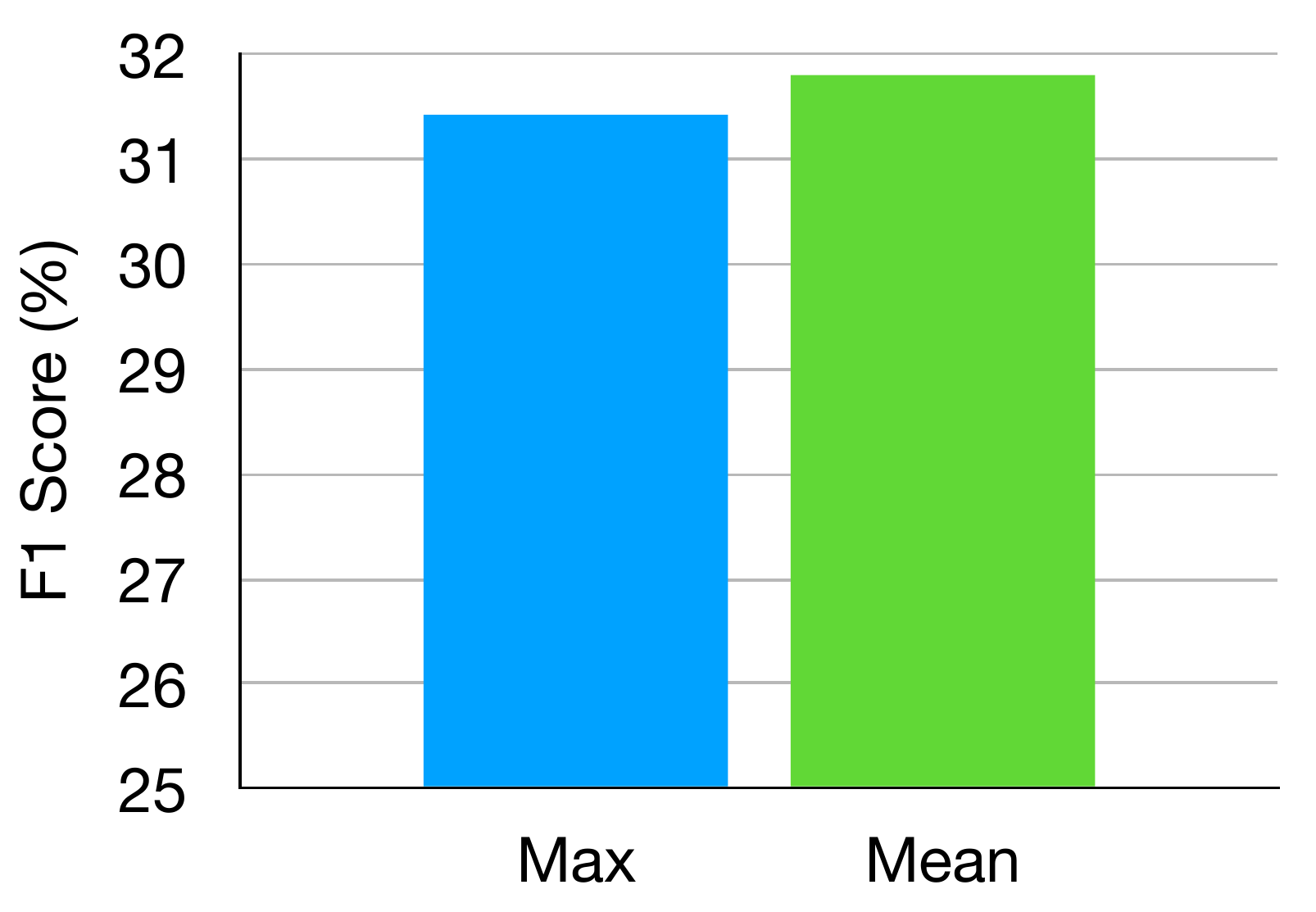}}%
\caption{{Parameter study with BERT$_{\rm BASE}$ on TAC KBP-OIE hold-out subset.}}
\label{fig:para}
\end{figure*}

\subsection{Error Analysis}
\label{sec:error}
There is still significant room to improve the results. We argue that we are measuring a lower bound for what LMs know. To further understand the shortcomings of the current method, we conduct an error analysis of the errors in precision on all datasets. We choose BERT$_{\rm LARGE}$ for the study. We sample 100 documents from the Wikidata-OIE dataset, and manually check the reasons for the errors. We find {33.1\%} of the errors are caused by {\em incorrect arguments}, while the predicate phrases are correct. The errors are due to the incorrect noun chunks detected by the spaCy. {18.3\%} of the errors are due to the {\em missing pairs in predicate mapping}. We also note that approximately {23.8\%} of the errors are actually {\em correct triples that are not covered by Wikidata}. For example, {\sl (Bob\_Dylan, residence, Nashville)} does not exist in Wikidata, but it is a correct triple. The rest of the errors made by BERT$_{\rm LARGE}$ are {\em incorrect predicate phrases}, such as uninformative phrases. We find similar errors are made by BERT$_{\rm LARGE}$ on other datasets. Based on the above analysis, enhancing argument detection and predicate mapping is helpful to further improve the results.

\subsection{Runtime Analysis}
\label{sec:runtime}
The runtime of OIE systems is crucial in practice. We test the runtime of different OIE systems on Re-OIE2016. The results are in Table~\ref{tab:runtime}. We find ours is competitive in terms of efficiency given the size of the models.

\subsection{Parameter Study}
\label{sec:para}
We study the effects of the key parameters using BERT$_{\rm BASE}$ on TAC KBP-OIE as shown in Figure~\ref{fig:para}. We randomly sample 20\% of the oracle query entities (provided by TAC KBP) as a hold-out dataset to tune the parameters, and use the best parameter setting achieved for all experiments. When studying the effect of a certain parameter, we keep the remaining parameters as default. We use F1 to measure the effects. Additional details are described in Appendix~\ref{sec:appendixparaset}.

\section{Related Work}
Pre-trained language models (LM), e.g., BERT~\citep{devlin2018bert}, GPT~\citep{radford2018improving,radford2019language}, and large LMs over 100B parameters~\citep{brown2020language,chowdhery2022palm,zeng2022glm} contain growing amount of linguistic and factual knowledge obtained via pre-training on large-scale corpora. To evaluate their abilities, researchers have created many knowledge benchmarks. LAMA leverages manually created prompts~\citep{petroni2019language,petroni2020context}. Recent studies have also developed soft prompts~\cite{liu2021gpt,zhong2021factual}) for fact retrieval. KILT~\citep{petroni2020kilt} proposes a knowledge-intensive benchmark concerning several downstream tasks to evaluate LMs' ability in capturing knowledge. 
\citet{wang-etal-2022-deepstruct} have utilized a set of knowledge-intensive structure prediction tasks to evaluate the knowledge in pre-trained LMs.
\citet{shen-etal-2022-lass} have adapted KG completion as a benchmark to evaluate LMs.
Besides relational knowledge, closed-book OpenQA~\citep{roberts2020much} benchmarks (in which LMs answer the open-domain questions without retrieving contexts)
have also been adopted as a way to evaluate LMs' knowledge. 
While the existing benchmarks evaluate LMs in an implicit way, the main difference is that our benchmark explicitly and interpretably evaluates triples from the textual corpora extracted using model parameters (e.g. attentions).
In the field of neural network interpretation~\citep{linzen2016assessing,adi2016fine,tenney2019you}, in particular the pre-trained deep LM analysis, substantial recent work focuses on both visualizing and analyzing the attention~\citep{vig2019visualizing,jain2019attention,clark2019does,michel2019sixteen,ramsauer2020hopfield}. Instead of analyzing or visualizing, our benchmark quantitatively evaluates the relational information with respect to open information extraction.

Open information extraction systems, e.g., OLLIE~\citep{schmitz2012open}, Reverb~\citep{fader2011identifying}, Stanford OpenIE~\citep{angeli2015leveraging}, OpenIE 5~\citep{saha2017bootstrapping,saha2018open}, RnnOIE~\citep{stanovsky2018supervised}, and OpenIE 6~\citep{kolluru2020openie6} aim to extract triples from web corpora for open schema KGs. Besides, NELL~\citep{carlson2010toward}, DeepDive~\citep{niu2012elementary}, Knowledge Vault~\citep{dong2014knowledge} extract information based on a fixed schema or ontology, where humans help improve the accuracy of the extractions. Probase~\citep{wu2012probase} produces taxonomies instead of rich typed relations in general KGs. Our benchmark first evaluates LMs' unsupervised information extraction ability on common open information extraction datasets such as CaRB~\cite{bhardwaj2019carb} and Re-OIE2016~\cite{zhan2020span}, and then aligns the extracted triples to KG triples for large-scale knowledge extraction benchmark construction. Our algorithm is similar to the generation algorithm in DeepEx~\cite{wang2021zero}. The focus of this work is to benchmark the zero-shot OIE performance of pre-trained LMs on both standard and factual OIE datasets. To further improve the OIE performance, the ranking module in DeepEx can be useful. The structure pre-training proposed in ~\cite{wang-etal-2022-deepstruct} can also be helpful.

\section{Conclusion}

We benchmark the general relational information in pre-trained language models (LM) in an open information extraction (OIE) setup. We find that the pre-trained LMs contain a considerable amount of open relational information through large-scale evaluation on both standard OIE datasets and newly created large-scale factual OIE datasets in our \benchmark\ benchmark. We are able to turn pre-training LMs into zero-shot OIE systems to efficiently deliver the benchmark results. The reach of this result is broad and has potential downstream utility for deep neural network interpretation, information extraction, and knowledge graph construction. Although the results are promising, we argue that our results just indicate a lower bound about what the LMs have. We hope our results will foster further research in the LM OIE benchmark direction.

\section{Limitations}
For the limitations of our method, the argument extraction module of our algorithm relies on a third-party noun chunker. As reported, the noun chunker introduces the majority of the errors in our extraction results. A limitation in our benchmark is that we have not conducted a large-scale manual evaluation of our factual OIE datasets (TAC KBP-OIE and Wikidata-OIE). The main focus of our study is to provide a large-scale OIE benchmark. As a result, this makes our benchmark more challenging to be used than standard OIE datasets in terms of computation costs and infrastructure. Finally, we have only benchmarked BERT and GPT-2 on our datasets. Future work could include testing a wide range of language models on our benchmark.

\section{Ethical Considerations}
We hereby acknowledge that all of the co-authors of this work are aware of the provided \textit{ACM Code of Ethics} and honor the code of conduct. This work is about benchmarking the zero-shot OIE capability of pre-trained language models including BERT and GPT. Our ethical considerations and the work's underlying future impacts are discussed in the following perspectives. Language models are known to present potential risks and limitations~\cite{brown2020language}, and the corpus used in pre-training (such as Wikipedia) may introduce unwanted biases and toxicity. We do not anticipate the production of harmful outputs after using our method or datasets, especially for vulnerable populations.

\section{Environmental Impact}
We adopt the pre-trained language models BERT~\citep{devlin2018bert} and GPT-2 series~\citep{radford2019language} in our IELM benchmark. The models' carbon footprints are estimated to be 22–28 kilograms~\citep{gibney2022shrink}. Additionally, The focus of this study is to test the zero-shot OIE ability of pre-trained language models. We do not train language models on massive datasets. Instead, we only do inference on a few evaluation datasets. This cost is less than 0.1\% energy than that of their pre-training. This demonstrates that developing proper zero-shot learning strategies for large language models can not only deepen our understanding of their latent mechanisms, but also further reduce the energy consumption and environmental impacts that language models with ever-growing size may cause.

\section*{Acknowledgements}
We would like to thank the anonymous reviewers for their suggestions and comments. This material is in part based upon work supported by Berkeley DeepDrive and Berkeley Artificial Intelligence Research.

\bibliography{custom}

\begin{thebibliography}{56}
\expandafter\ifx\csname natexlab\endcsname\relax\def\natexlab#1{#1}\fi

\bibitem[{Adi et~al.(2016)Adi, Kermany, Belinkov, Lavi, and
  Goldberg}]{adi2016fine}
Yossi Adi, Einat Kermany, Yonatan Belinkov, Ofer Lavi, and Yoav Goldberg. 2016.
\newblock Fine-grained analysis of sentence embeddings using auxiliary
  prediction tasks.
\newblock \emph{arXiv preprint arXiv:1608.04207}.

\bibitem[{Angeli et~al.(2015)Angeli, Premkumar, and
  Manning}]{angeli2015leveraging}
Gabor Angeli, Melvin Jose~Johnson Premkumar, and Christopher~D Manning. 2015.
\newblock Leveraging linguistic structure for open domain information
  extraction.
\newblock In \emph{ACL}, pages 344--354.

\bibitem[{Bhardwaj et~al.(2019)Bhardwaj, Aggarwal, and
  Mausam}]{bhardwaj2019carb}
Sangnie Bhardwaj, Samarth Aggarwal, and Mausam Mausam. 2019.
\newblock Carb: A crowdsourced benchmark for open ie.
\newblock In \emph{EMNLP}, pages 6262--6267.

\bibitem[{Brown et~al.(2020)Brown, Mann, Ryder, Subbiah, Kaplan, Dhariwal,
  Neelakantan, Shyam, Sastry, Askell et~al.}]{brown2020language}
Tom~B Brown, Benjamin Mann, Nick Ryder, Melanie Subbiah, Jared Kaplan, Prafulla
  Dhariwal, Arvind Neelakantan, Pranav Shyam, Girish Sastry, Amanda Askell,
  et~al. 2020.
\newblock Language models are few-shot learners.
\newblock \emph{arXiv preprint arXiv:2005.14165}.

\bibitem[{Carlson et~al.(2010)Carlson, Betteridge, Kisiel, Settles,
  Hruschka~Jr, and Mitchell}]{carlson2010toward}
Andrew Carlson, Justin Betteridge, Bryan Kisiel, Burr Settles, Estevam~R
  Hruschka~Jr, and Tom~M Mitchell. 2010.
\newblock Toward an architecture for never-ending language learning.
\newblock In \emph{AAAI}, 3.

\bibitem[{Chowdhery et~al.(2022)Chowdhery, Narang, Devlin, Bosma, Mishra,
  Roberts, Barham, Chung, Sutton, Gehrmann et~al.}]{chowdhery2022palm}
Aakanksha Chowdhery, Sharan Narang, Jacob Devlin, Maarten Bosma, Gaurav Mishra,
  Adam Roberts, Paul Barham, Hyung~Won Chung, Charles Sutton, Sebastian
  Gehrmann, et~al. 2022.
\newblock Palm: Scaling language modeling with pathways.
\newblock \emph{arXiv preprint arXiv:2204.02311}.

\bibitem[{Christensen et~al.(2011)Christensen, Soderland, and
  Etzioni}]{christensen2011analysis}
Janara Christensen, Stephen Soderland, and Oren Etzioni. 2011.
\newblock An analysis of open information extraction based on semantic role
  labeling.
\newblock In \emph{KCAP}, pages 113--120.

\bibitem[{Clark et~al.(2019)Clark, Khandelwal, Levy, and
  Manning}]{clark2019does}
Kevin Clark, Urvashi Khandelwal, Omer Levy, and Christopher~D Manning. 2019.
\newblock What does bert look at? an analysis of bert's attention.
\newblock \emph{arXiv preprint arXiv:1906.04341}.

\bibitem[{Cui et~al.(2018)Cui, Wei, and Zhou}]{cui2018neural}
Lei Cui, Furu Wei, and Ming Zhou. 2018.
\newblock Neural open information extraction.
\newblock In \emph{ACL}, pages 407--413.

\bibitem[{Del~Corro and Gemulla(2013)}]{del2013clausie}
Luciano Del~Corro and Rainer Gemulla. 2013.
\newblock Clausie: clause-based open information extraction.
\newblock In \emph{WWW}, pages 355--366.

\bibitem[{Devlin et~al.(2018)Devlin, Chang, Lee, and
  Toutanova}]{devlin2018bert}
Jacob Devlin, Ming-Wei Chang, Kenton Lee, and Kristina Toutanova. 2018.
\newblock Bert: Pre-training of deep bidirectional transformers for language
  understanding.
\newblock \emph{arXiv preprint arXiv:1810.04805}.

\bibitem[{Dong et~al.(2014)Dong, Gabrilovich, Heitz, Horn, Lao, Murphy,
  Strohmann, Sun, and Zhang}]{dong2014knowledge}
Xin Dong, Evgeniy Gabrilovich, Geremy Heitz, Wilko Horn, Ni~Lao, Kevin Murphy,
  Thomas Strohmann, Shaohua Sun, and Wei Zhang. 2014.
\newblock Knowledge vault: A web-scale approach to probabilistic knowledge
  fusion.
\newblock In \emph{KDD}, pages 601--610.

\bibitem[{Fader et~al.(2011)Fader, Soderland, and
  Etzioni}]{fader2011identifying}
Anthony Fader, Stephen Soderland, and Oren Etzioni. 2011.
\newblock Identifying relations for open information extraction.
\newblock In \emph{EMNLP}, pages 1535--1545.

\bibitem[{Gashteovski et~al.(2017)Gashteovski, Gemulla, and
  Corro}]{gashteovski2017minie}
Kiril Gashteovski, Rainer Gemulla, and Luciano~del Corro. 2017.
\newblock Minie: minimizing facts in open information extraction.
\newblock ACL.

\bibitem[{Gibney(2022)}]{gibney2022shrink}
Elizabeth Gibney. 2022.
\newblock How to shrink ai’s ballooning carbon footprint.
\newblock \emph{Nature}, 607(7920):648--648.

\bibitem[{Hewitt and Manning(2019)}]{hewitt2019structural}
John Hewitt and Christopher~D Manning. 2019.
\newblock A structural probe for finding syntax in word representations.
\newblock In \emph{NAACL}, pages 4129--4138.

\bibitem[{Hoffart et~al.(2011)Hoffart, Yosef, Bordino, F{\"u}rstenau, Pinkal,
  Spaniol, Taneva, Thater, and Weikum}]{hoffart2011robust}
Johannes Hoffart, Mohamed~Amir Yosef, Ilaria Bordino, Hagen F{\"u}rstenau,
  Manfred Pinkal, Marc Spaniol, Bilyana Taneva, Stefan Thater, and Gerhard
  Weikum. 2011.
\newblock Robust disambiguation of named entities in text.
\newblock In \emph{EMNLP}, pages 782--792.

\bibitem[{Jain and Wallace(2019)}]{jain2019attention}
Sarthak Jain and Byron~C Wallace. 2019.
\newblock Attention is not explanation.
\newblock \emph{arXiv preprint arXiv:1902.10186}.

\bibitem[{Jiang et~al.(2020)Jiang, Xu, Araki, and Neubig}]{jiang2020can}
Zhengbao Jiang, Frank~F Xu, Jun Araki, and Graham Neubig. 2020.
\newblock How can we know what language models know?
\newblock \emph{TACL}, 8:423--438.

\bibitem[{Kolitsas et~al.(2018)Kolitsas, Ganea, and Hofmann}]{kolitsas2018end}
Nikolaos Kolitsas, Octavian-Eugen Ganea, and Thomas Hofmann. 2018.
\newblock End-to-end neural entity linking.
\newblock \emph{arXiv preprint arXiv:1808.07699}.

\bibitem[{Kolluru et~al.(2020{\natexlab{a}})Kolluru, Adlakha, Aggarwal,
  Chakrabarti et~al.}]{kolluru2020openie6}
Keshav Kolluru, Vaibhav Adlakha, Samarth Aggarwal, Soumen Chakrabarti, et~al.
  2020{\natexlab{a}}.
\newblock Openie6: Iterative grid labeling and coordination analysis for open
  information extraction.
\newblock In \emph{EMNLP}, pages 3748--3761.

\bibitem[{Kolluru et~al.(2020{\natexlab{b}})Kolluru, Aggarwal, Rathore,
  Chakrabarti et~al.}]{kolluru2020imojie}
Keshav Kolluru, Samarth Aggarwal, Vipul Rathore, Soumen Chakrabarti, et~al.
  2020{\natexlab{b}}.
\newblock Imojie: Iterative memory-based joint open information extraction.
\newblock In \emph{ACL}, pages 5871--5886.

\bibitem[{Linzen et~al.(2016)Linzen, Dupoux, and
  Goldberg}]{linzen2016assessing}
Tal Linzen, Emmanuel Dupoux, and Yoav Goldberg. 2016.
\newblock Assessing the ability of lstms to learn syntax-sensitive
  dependencies.
\newblock \emph{TACL}, pages 521--535.

\bibitem[{Liu et~al.(2021)Liu, Zheng, Du, Ding, Qian, Yang, and
  Tang}]{liu2021gpt}
Xiao Liu, Yanan Zheng, Zhengxiao Du, Ming Ding, Yujie Qian, Zhilin Yang, and
  Jie Tang. 2021.
\newblock Gpt understands, too.
\newblock \emph{arXiv:2103.10385}.

\bibitem[{Michel et~al.(2019)Michel, Levy, and Neubig}]{michel2019sixteen}
Paul Michel, Omer Levy, and Graham Neubig. 2019.
\newblock Are sixteen heads really better than one?
\newblock In \emph{NIPS}, pages 14014--14024.

\bibitem[{Niu et~al.(2012)Niu, Zhang, R{\'e}, and Shavlik}]{niu2012elementary}
Feng Niu, Ce~Zhang, Christopher R{\'e}, and Jude Shavlik. 2012.
\newblock Elementary: Large-scale knowledge-base construction via machine
  learning and statistical inference.
\newblock \emph{International Journal on Semantic Web and Information Systems
  (IJSWIS)}, 8(3):42--73.

\bibitem[{Petroni et~al.(2020)Petroni, Lewis, Piktus, Rockt{\"a}schel, Wu,
  Miller, and Riedel}]{petroni2020context}
Fabio Petroni, Patrick Lewis, Aleksandra Piktus, Tim Rockt{\"a}schel, Yuxiang
  Wu, Alexander~H Miller, and Sebastian Riedel. 2020.
\newblock How context affects language models' factual predictions.
\newblock \emph{arXiv preprint arXiv:2005.04611}.

\bibitem[{Petroni et~al.(2021)Petroni, Piktus, Fan, Lewis, Yazdani, De~Cao,
  Thorne, Jernite, Karpukhin, Maillard et~al.}]{petroni2020kilt}
Fabio Petroni, Aleksandra Piktus, Angela Fan, Patrick Lewis, Majid Yazdani,
  Nicola De~Cao, James Thorne, Yacine Jernite, Vladimir Karpukhin, Jean
  Maillard, et~al. 2021.
\newblock Kilt: a benchmark for knowledge intensive language tasks.
\newblock In \emph{NAACL}, pages 2523--2544.

\bibitem[{Petroni et~al.(2019)Petroni, Rockt{\"a}schel, Lewis, Bakhtin, Wu,
  Miller, and Riedel}]{petroni2019language}
Fabio Petroni, Tim Rockt{\"a}schel, Patrick Lewis, Anton Bakhtin, Yuxiang Wu,
  Alexander~H Miller, and Sebastian Riedel. 2019.
\newblock Language models as knowledge bases?
\newblock \emph{arXiv preprint arXiv:1909.01066}.

\bibitem[{Radford et~al.(2018)Radford, Narasimhan, Salimans, and
  Sutskever}]{radford2018improving}
Alec Radford, Karthik Narasimhan, Tim Salimans, and Ilya Sutskever. 2018.
\newblock Improving language understanding by generative pre-training.

\bibitem[{Radford et~al.(2019)Radford, Wu, Child, Luan, Amodei, and
  Sutskever}]{radford2019language}
Alec Radford, Jeffrey Wu, Rewon Child, David Luan, Dario Amodei, and Ilya
  Sutskever. 2019.
\newblock Language models are unsupervised multitask learners.
\newblock \emph{OpenAI Blog}, (8):9.

\bibitem[{Raffel et~al.(2019)Raffel, Shazeer, Roberts, Lee, Narang, Matena,
  Zhou, Li, and Liu}]{abs-1910-10683}
Colin Raffel, Noam Shazeer, Adam Roberts, Katherine Lee, Sharan Narang, Michael
  Matena, Yanqi Zhou, Wei Li, and Peter~J. Liu. 2019.
\newblock Exploring the limits of transfer learning with a unified text-to-text
  transformer.
\newblock \emph{CoRR}, abs/1910.10683.

\bibitem[{Ramsauer et~al.(2020)Ramsauer, Sch{\"a}fl, Lehner, Seidl, Widrich,
  Gruber, Holzleitner, Pavlovi{\'c}, Sandve, Greiff
  et~al.}]{ramsauer2020hopfield}
Hubert Ramsauer, Bernhard Sch{\"a}fl, Johannes Lehner, Philipp Seidl, Michael
  Widrich, Lukas Gruber, Markus Holzleitner, Milena Pavlovi{\'c}, Geir~Kjetil
  Sandve, Victor Greiff, et~al. 2020.
\newblock Hopfield networks is all you need.
\newblock \emph{arXiv preprint arXiv:2008.02217}.

\bibitem[{Ro et~al.(2020)Ro, Lee, and Kang}]{ro2020multi}
Youngbin Ro, Yukyung Lee, and Pilsung Kang. 2020.
\newblock Multiˆ2oie: Multilingual open information extraction based on
  multi-head attention with bert.
\newblock In \emph{Findings of EMNLP}, pages 1107--1117.

\bibitem[{Roberts et~al.(2020)Roberts, Raffel, and Shazeer}]{roberts2020much}
Adam Roberts, Colin Raffel, and Noam Shazeer. 2020.
\newblock How much knowledge can you pack into the parameters of a language
  model?
\newblock \emph{arXiv preprint arXiv:2002.08910}.

\bibitem[{Roy et~al.(2019)Roy, Park, Lee, and Pan}]{roy2019supervising}
Arpita Roy, Youngja Park, Taesung Lee, and Shimei Pan. 2019.
\newblock Supervising unsupervised open information extraction models.
\newblock In \emph{EMNLP}, pages 728--737.

\bibitem[{Saha et~al.(2017)Saha, Pal et~al.}]{saha2017bootstrapping}
Swarnadeep Saha, Harinder Pal, et~al. 2017.
\newblock Bootstrapping for numerical open ie.
\newblock In \emph{ACL}, pages 317--323.

\bibitem[{Saha et~al.(2018)}]{saha2018open}
Swarnadeep Saha et~al. 2018.
\newblock Open information extraction from conjunctive sentences.
\newblock In \emph{COLING}, pages 2288--2299.

\bibitem[{Schmitz et~al.(2012)Schmitz, Soderland, Bart, Etzioni
  et~al.}]{schmitz2012open}
Michael Schmitz, Stephen Soderland, Robert Bart, Oren Etzioni, et~al. 2012.
\newblock Open language learning for information extraction.
\newblock In \emph{EMNLP}, pages 523--534.

\bibitem[{Shen et~al.(2022)Shen, Wang, Gong, and Song}]{shen-etal-2022-lass}
Jianhao Shen, Chenguang Wang, Linyuan Gong, and Dawn Song. 2022.
\newblock Joint language semantic and structure embedding for knowledge graph
  completion.
\newblock In \emph{COLING}.

\bibitem[{Spitkovsky and Chang(2012)}]{spitkovsky2012cross}
Valentin~I Spitkovsky and Angel~X Chang. 2012.
\newblock A cross-lingual dictionary for english wikipedia concepts.

\bibitem[{Stanovsky and Dagan(2016)}]{stanovsky2016creating}
Gabriel Stanovsky and Ido Dagan. 2016.
\newblock Creating a large benchmark for open information extraction.
\newblock In \emph{EMNLP}, pages 2300--2305.

\bibitem[{Stanovsky et~al.(2016)Stanovsky, Ficler, Dagan, and
  Goldberg}]{stanovsky2016getting}
Gabriel Stanovsky, Jessica Ficler, Ido Dagan, and Yoav Goldberg. 2016.
\newblock Getting more out of syntax with props.
\newblock \emph{arXiv preprint arXiv:1603.01648}.

\bibitem[{Stanovsky et~al.(2018)Stanovsky, Michael, Zettlemoyer, and
  Dagan}]{stanovsky2018supervised}
Gabriel Stanovsky, Julian Michael, Luke Zettlemoyer, and Ido Dagan. 2018.
\newblock Supervised open information extraction.
\newblock In \emph{NAACL}, pages 885--895.

\bibitem[{Surdeanu(2013)}]{surdeanu2013overview}
Mihai Surdeanu. 2013.
\newblock Overview of the tac2013 knowledge base population evaluation: English
  slot filling and temporal slot filling.
\newblock \emph{TAC}, page~2.

\bibitem[{Talmor et~al.(2019)Talmor, Elazar, Goldberg, and
  Berant}]{talmor2019olmpics}
Alon Talmor, Yanai Elazar, Yoav Goldberg, and Jonathan Berant. 2019.
\newblock olmpics--on what language model pre-training captures.
\newblock \emph{arXiv preprint arXiv:1912.13283}.

\bibitem[{Tenney et~al.(2019)Tenney, Xia, Chen, Wang, Poliak, McCoy, Kim,
  Van~Durme, Bowman, Das et~al.}]{tenney2019you}
Ian Tenney, Patrick Xia, Berlin Chen, Alex Wang, Adam Poliak, R~Thomas McCoy,
  Najoung Kim, Benjamin Van~Durme, Samuel~R Bowman, Dipanjan Das, et~al. 2019.
\newblock What do you learn from context? probing for sentence structure in
  contextualized word representations.
\newblock \emph{arXiv preprint arXiv:1905.06316}.

\bibitem[{Vaswani et~al.(2017)Vaswani, Shazeer, Parmar, Uszkoreit, Jones,
  Gomez, Kaiser, and Polosukhin}]{vaswani2017attention}
Ashish Vaswani, Noam Shazeer, Niki Parmar, Jakob Uszkoreit, Llion Jones,
  Aidan~N Gomez, {\L}ukasz Kaiser, and Illia Polosukhin. 2017.
\newblock Attention is all you need.
\newblock In \emph{NIPS}, pages 5998--6008.

\bibitem[{Vig(2019)}]{vig2019visualizing}
Jesse Vig. 2019.
\newblock Visualizing attention in transformerbased language models.
\newblock \emph{arXiv preprint arXiv:1904.02679}.

\bibitem[{Wang et~al.(2021)Wang, Liu, Chen, Hong, Tang, and
  Song}]{wang2021zero}
Chenguang Wang, Xiao Liu, Zui Chen, Haoyun Hong, Jie Tang, and Dawn Song. 2021.
\newblock Zero-shot information extraction as a unified text-to-triple
  translation.
\newblock In \emph{EMNLP}.

\bibitem[{Wang et~al.(2022)Wang, Liu, Chen, Hong, Tang, and
  Song}]{wang-etal-2022-deepstruct}
Chenguang Wang, Xiao Liu, Zui Chen, Haoyun Hong, Jie Tang, and Dawn Song. 2022.
\newblock {D}eep{S}truct: Pretraining of language models for structure
  prediction.
\newblock In \emph{ACL}.

\bibitem[{Wu et~al.(2012)Wu, Li, Wang, and Zhu}]{wu2012probase}
Wentao Wu, Hongsong Li, Haixun Wang, and Kenny~Q Zhu. 2012.
\newblock Probase: A probabilistic taxonomy for text understanding.
\newblock In \emph{SIGMOD}, pages 481--492.

\bibitem[{Zeng et~al.(2022)Zeng, Liu, Du, Wang, Lai, Ding, Yang, Xu, Zheng, Xia
  et~al.}]{zeng2022glm}
Aohan Zeng, Xiao Liu, Zhengxiao Du, Zihan Wang, Hanyu Lai, Ming Ding, Zhuoyi
  Yang, Yifan Xu, Wendi Zheng, Xiao Xia, et~al. 2022.
\newblock Glm-130b: An open bilingual pre-trained model.
\newblock \emph{arXiv preprint arXiv:2210.02414}.

\bibitem[{Zhan and Zhao(2020)}]{zhan2020span}
Junlang Zhan and Hai Zhao. 2020.
\newblock Span model for open information extraction on accurate corpus.
\newblock In \emph{AAAI}, pages 9523--9530.

\bibitem[{Zhang et~al.(2017)Zhang, Zhong, Chen, Angeli, and
  Manning}]{zhang2017position}
Yuhao Zhang, Victor Zhong, Danqi Chen, Gabor Angeli, and Christopher~D Manning.
  2017.
\newblock Position-aware attention and supervised data improve slot filling.
\newblock In \emph{EMNLP}.

\bibitem[{Zhong et~al.(2021)Zhong, Friedman, and Chen}]{zhong2021factual}
Zexuan Zhong, Dan Friedman, and Danqi Chen. 2021.
\newblock Factual probing is [mask]: Learning vs. learning to recall.
\newblock In \emph{NAACL}, pages 5017--5033.

\end{thebibliography}
\bibliographystyle{acl_natbib}

\appendix
\section{The \benchmark\ Benchmark Details}
\label{sec:adddetails}
Additional details of our open information extraction (OIE) benchmark \benchmark\ are described in this section.

\subsection{Wikidata-OIE}
In this section, we describe some technical details regarding the construction and evaluation of Wikidata-OIE.

\subsubsection{Entity Linking}
We use an unsupervised entity linker for both Wikidata-OIE dataset construction and OIE evaluation. The entity linker is originally developed in \citep{spitkovsky2012cross}, which is based on a mention-to-entity dictionary. We build an enhanced dictionary as follows: we add new Wikipedia anchors to the dictionary which results in 26 million entries compared to the original 21 million entries. Then a Wikipedia anchor to the Wikidata item dictionary is used to further link the entities (or arguments) to Wikidata. If an argument is a pronoun, we further use neuralcoref~\footnote{\tiny\url{https://github.com/huggingface/neuralcoref}} for coreference resolution.

\subsubsection{Predicate Mapping}
The predicate mapping of Wikidata-OIE is constructed offline using the method in Sec.~\ref{sec:evalmethod}. In more detail, we randomly sampled a hold-out dataset including 2,000 documents from English Wikipedia for the bootstrapped predicate mapping construction based on the TAC KBP mapping~\citep{angeli2015leveraging}. To filter out the wrong predicate pairs, we manually check whether the top predicate phrases are true. 

\subsubsection{Gold Triples}
For gold triples in Wikidata-OIE, we only preserve those triples describing predicates between arguments that can be linked to corresponding Wikipedia anchors. We rule out triples of attributes about arguments and triples of auxiliary predicates (such as \textit{topic's main category.P901}) and finally result in 27,368,562 gold triple extractions.

\subsubsection{Evaluation}
Given the large number of source sentences and gold triples in Wikidata-OIE, a MongoDB database is maintained to store the gold triples to enable an efficient evaluation.

\subsection{Zero-Shot Language Model Based Open Information Extraction}
In this section, we introduce additional details about how we adapt pre-trained language models (LM) as zero-shot OIE systems.

\subsubsection{Argument Extraction}
We use spaCy noun chunker~\footnote{\tiny{\url{https://spacy.io/usage/linguistic-features/\#noun-chunks}}} to annotate the noun phrases in the sentences.

\subsubsection{Predicate Extraction}
We first describe predicate extraction introduced in Sec.~\ref{sec:predicate} in detail.

\begin{itemize}[leftmargin=*]
\item {\bf Beam Search}. The inputs of the search algorithm are an argument pair $(arg_0,arg_1)$, a sentence $s$, an attention matrix $\mathbf{A}_s$ of $s$. Both $arg_0$ and $arg_1$ are identified by the noun chunker in $s$. $\mathbf{A}_s$ is the attention matrix associated with $s$ from the forward pass of an LM without fine-tuning. The search gets started by adding the first argument $arg_0$ as the initial candidate in the beam. While there are still new candidates waiting to be yielded, the search continues, and the top $k$ candidates sorted by the attention scores are maintained in the beam.
The details of the proposed beam search are described in Algorithm~\ref{alg:beamsearch}. In practice, we implement an action manager $\mathcal{O}$ to decide which action to take at each step. Given a candidate $c$ in the beam, $\mathcal{O}(c) = \texttu{start}$ always happens at the beginning of the search. If $c$ has not reached the second argument $arg_1$ yet, $\mathcal{O}(c) = \mathbf{\texttu{yield}}$. Otherwise, $\mathcal{O}(c) = \texttu{stop}$.

\begin{algorithm}[tb]
\small
\caption{\label{alg:beamsearch}{Beam search with attention scores.}}
\begin{algorithmic}[1]
\Require{Argument pair $(arg_0,arg_1)$, sentence $s$, attention matrix $\mathbf{A_s}$, action manager $\mathcal{O}=\{\texttu{start}, \texttu{yield}, \texttu{stop}\}$, beam size $k$}
\Ensure{Triples $\mathbb{T}_{(arg_0,arg_1)}$}
\State $\mathbb{T}_{(arg_0,arg_1)} \leftarrow \{\texttu{start}(arg_0)\}$ 
\Comment{{\em Start} by adding the first argument as a candidate in the beam}
\While{$\exists c \in \mathbb{T}_{(arg_0,arg_1)} [\mathcal{O}(c) = \texttu{yield}]$} 
\State $\widetilde{\mathbb{T}}_{(arg_0,arg_1)} \leftarrow \emptyset$ 
\Comment{Initialize a new beam}
\ForAll{$c \in \mathbb{T}_{(arg_0,arg_1)}$}
\If{$\mathcal{O}(c) = \texttu{yield}$} 
\State $\widetilde{\mathbb{T}}_{(arg_0,arg_1)} \leftarrow  \widetilde{\mathbb{T}}_{(arg_0,arg_1)} \cup \{\texttu{yield}(c,s,\mathbf{A}_s)\}$ \Comment{{\em Yield} a new candidate if not reached the second argument}
\Else
\State $\widetilde{\mathbb{T}}_{(arg_0,arg_1)} \leftarrow  \widetilde{\mathbb{T}}_{(arg_0,arg_1)} \cup \{\texttu{stop}(c,t)\}$ \Comment{{\em Stop} then produce a valid triple if reached the second argument}
\EndIf
\EndFor
\State $\mathbb{T}_{(arg_0,arg_1)} \leftarrow {\rm TOP}(k, \widetilde{\mathbb{T}}_{(arg_0,arg_1)})$ \Comment{Maintain $k$-best candidates in the beam}
\EndWhile
\State \Return $\mathbb{T}_{(arg_0,arg_1)}$
\end{algorithmic}
\end{algorithm}

\item {\bf Implementation Details}. For Wikidata-OIE, we randomly split the English Wikipedia data into 20 partitions, and map the data partitions to 20 distributed servers to run. Each server is configured with four Tesla K80 12Gs. We set the max sequence length to 256, and batch size as 32 for BERT$_{\rm LARGE}$ and 4 for GPT-2$_{\rm XL}$. We use implementations of pre-trained LMs in the Transformers package~\footnote{\tiny\url{https://github.com/huggingface/transformers}}. We use spaCy sentencizer~\footnote{{\tiny\url{https://spacy.io/api/sentencizer}}} to segment the documents into sentences. BERT$_{\rm LARGE}$ takes approximately 48 hours, and GPT-2$_{\rm XL}$ costs around 96 hours. The resulting triples from the 20 servers are then reduced to a data server. The batch sizes of BERT$_{\rm BASE}$, GPT-2, GPT-2$_{\rm MEDIUM}$, GPT-2$_{\rm LARGE}$ are 64, 32, 16, 8 respectively.
\end{itemize}

\subsubsection{Parameter Settings}
\label{sec:appendixparaset}
We then discuss the parameter setup of our OIE systems as below.

The parameter settings are shared across all OIE datasets. All the choices are based on the parameter study in Sec.~\ref{sec:para}. The beam size of Algorithm~\ref{alg:beamsearch} is set to 6. The attention score threshold is set to 0.005, and the number of relation/predicate frequencies is set to 10. To generate the attention weight matrix $\mathbf{A}_s$ of a sentence, we reduce the weights of every attention head in the last layer of pre-trained LMs using the mean operator. We analyze the effects of various parameters below.

Figure~\ref{fig:para}(a) illustrates the effects of various beam sizes in Algorithm~\ref{alg:beamsearch}. We find that in general, the larger the beam size is, the better F1 the setting achieves. This is because our method is able to reserve more potentially correct triples when more candidates are allowed. However, F1 improvement gradually becomes subtle, while the computation costs increase more significantly. For efficiency consideration, we do not explore larger beam sizes. We set the beam size as 6.

Figure~\ref{fig:para}(b) compares the effect of different thresholds of the total score. We set the threshold as 0.005 since it achieves the best result. Note that the summed attention score is normalized by the length of the triple to penalize the cumbersome triples. The threshold is effective. This is mainly because of the relational information contained in the self-attention matrix: the score in the attention matrix is representing the chance of the triples to be the true triples based on the stored information. Figure~\ref{fig:para}(c) shows the impact of the predicate frequency threshold in identifying common predicates. The best result is achieved when it equals 10. This shows that while our method mostly identifies frequent predicates, it is also able to capture some rare predicates.

Figure~\ref{fig:para}(d) shows the comparison between the attention weights of the last layer and the mean of all layers. The attention weights of the last layer perform better. This is due to the attention weights in lower layers being low-level linguistic knowledge according to \citep{clark2019does,ramsauer2020hopfield}, which are less relevant to the relational information. Figure~\ref{fig:para}(e) compares the impact of different attention reductions, i.e., mean, max, over the attention heads of the last layer. We find the ``mean'' performs better. The reason is that the token often intensively attends to several specific tokens in the sequence~\citep{michel2019sixteen}, and the ``mean'' operator is less sensitive to such biases.

\subsection{The Number of Predicates of Standard and Factual OIE Datasets}
\label{sec:appendixdatasetcomp}
Note that there are more predicates in standard OIE datasets than that in factual datasets. This is because, for standard OIE, predicates are open and not attached to a certain schema. These predicates were extracted from the input sentences and are usually natural language utterances. For factual OIE, the predicates are unified into a fixed KG schema. For example, for a person's birthplace, there are multiple natural language expressions like ``was born in'' or ``gave birth'' in standard OIE datasets, while only a single ``birth\_place'' predicate exists in the factual OIE sets.

\subsection{Predicate Mapping Examples}
\label{sec:appendixmapping}
We show example predicate mappings in a dictionary below.
\begin{itemize}
    \item \texttt{per:city\_of\_birth}: born in, born at, born, birth city, hometown.
    \item \texttt{org:founded\_by}: established by, founded by, founded, founder, co-founder of.
\end{itemize}
where the keys are KG predicates, e.g., \texttt{per:city\_of\_birth} and \texttt{org:founded\_by}. The values are the corresponding OIE relations.

\subsection{Comparison Systems}
\label{sec:appendixcompsys}
We compare our zero-shot OIE systems with the following OIE systems. 
\subsubsection{Neural OIE Systems}
The following neural network based systems are selected.
\begin{itemize}[leftmargin=*]
    \item SenseOIE~\citep{roy2019supervising}\footnote{\tiny{The code is not available.}} learns to ensemble various previous unsupervised OIE systems' extractions using supervised learning to combine their strengths.
    \item SpanOIE~\citep{zhan2020span}\footnote{\tiny\url{https://github.com/zhanjunlang/Span_OIE}} presents the Re-OIE2016 datasets for a more rigorous evaluation and a span-based (instead of sequence labeling) extraction model.
    \item RnnOIE~\citep{stanovsky2018supervised}\footnote{\tiny\url{https://github.com/gabrielStanovsky/supervised-oie}} is one of the state-of-the-art OIE systems. It uses LSTM to model the OIE problem as a sequence tagging problem, and is trained on a large-scale OIE training set.
    \item NeuralOIE~\citep{cui2018neural}\footnote{\tiny We use the BERT implementation available at \url{https://github.com/dair-iitd/imojie}.} is an encoder-decoder based architecture that adopts the copy mechanism to conduct OIE.
    \item IMOJIE~\citep{kolluru2020imojie}\footnote{\tiny\url{https://github.com/dair-iitd/imojie}} is a sequence generation based OIE model that uses BERT at encoding time.
    \item Multi$^2$OIE~\citep{ro2020multi}\footnote{\tiny\url{https://github.com/youngbin-ro/Multi2OIE}} models OIE as a sequence labeling problem that combines BERT with multi-head attention blocks.
    \item OpenIE6~\citep{kolluru2020openie6}\footnote{\tiny\url{https://github.com/dair-iitd/openie6}} is one of the state-of-the-art OIE systems. It treats OIE as a 2-D grid labeling task, and trains a BERT family architecture for the task.
\end{itemize}
Note that while our methods are zero-shot without needing to use the specific training sets, all the neural OIE systems are supervised on corresponding training sets.

\subsubsection{Linguistic OIE Systems}
We also compare our systems with the following linguistic pattern based systems developed prior to the use of neural networks.
\begin{itemize}[leftmargin=*]
    \item MinIE~\citep{gashteovski2017minie}\footnote{\tiny\url{https://github.com/uma-pi1/minie}} proposes to minimize facts in OIE by representing information by annotations rather than extraction and removing redundant specific information.
    \item ClausIE~\citep{del2013clausie}\footnote{\tiny\url{https://www.mpi-inf.mpg.de/departments/databases-and-information-systems/research/ambiverse-nlu/clausie}} is a clause-based approach by first identifying linguistic structure and then their information and attributes.
    \item OLLIE~\citep{schmitz2012open}\footnote{\tiny\url{https://github.com/knowitall/ollie}} uses contextual sentence decomposition to conduct OIE.
    \item PropS~\citep{stanovsky2016getting}\footnote{\tiny\url{https://github.com/gabrielStanovsky/props}} proposes proposition structure which is implied from syntax using dependency trees.
    \item OpenIE4~\citep{christensen2011analysis}\footnote{\tiny\url{https://github.com/allenai/openie-standalone}} is the successor to OLLIE using similar argument and relation expansion heuristics to create OIE extractions from semantic role labeling frames.
    \item OpenIE5~\citep{saha2017bootstrapping,saha2018open}\footnote{\label{ft:openie51}\tiny\url{https://github.com/dair-iitd/OpenIE-standalone}} is one of the state-of-the-art OIE systems, which is the successor to OLLIE, and it improves extractions from noun relations, numerical sentences, and conjunctive sentences depending on the linguistic patterns.
    \item Stanford OpenIE~\citep{angeli2015leveraging}\footnote{\tiny\url{https://nlp.stanford.edu/software/openie.html}} leverages POS tag and dependency parser, and generates self-contained clauses from long sentences to extract the triples.
\end{itemize}

\section{The TAC KBP-OIE and Wikidata-OIE Datasets}
We show samples of our zero-shot OIE extractions and the gold triples on both TAC KBP-OIE and Wikidata-OIE datasets.

\subsection{TAC KBP-OIE}
\paragraph{OIE Extractions and Gold Extractions} We randomly sample 100 documents from the TAC KBP-OIE corpus, then sample sentences from those documents. 
The uncurated triples and the corresponding gold triples of the sampled sentences based on our best methods
BERT$_{\rm LARGE}$ and GPT-2$_{\rm XL}$ are shown in Figure~\ref{fig:kbpmapbert} and Figure~\ref{fig:kbpmapgpt} respectively.
We also randomly sample sentences in which BERT$_{\rm LARGE}$ differs from GPT-2$_{\rm XL}$ in the resulting triples for comparison, which are illustrated in Figure~\ref{fig:kbpmapbertgpt}.
In each table, ``{\bf ID}'' represents the document ID of a sampled sentence in the TAC KBP-OIE corpus. ``{\bf Sentence}'' indicates the sampled sentence. ``{\bf Triples to gold triples}'' column contains the extraction triples (on the left side of ``$\rightarrow$'') and their corresponding gold triples (on the right side of ``$\rightarrow$'').

\subsection{Wikidata-OIE}

\paragraph{OIE Extractions and Gold Extractions} Similar to TAC KBP-OIE, we randomly sample 100 documents from the Wikidata-OIE corpus (i.e., English Wikipedia), then sample sentences from those documents. 
Similar to TAC KBP-OIE, Figure~\ref{fig:wikimapbert} and Figure~\ref{fig:wikimapgpt} show the uncurated triples and the corresponding gold triples of the sampled sentences based on our zero-shot systems BERT$_{\rm LARGE}$ and GPT-2$_{\rm XL}$ respectively.
Figure~\ref{fig:wikimapbertgpt} illustrates the randomly sampled sentences in which BERT$_{\rm LARGE}$ extracts different triples compared to that from GPT-2$_{\rm XL}$.
In each table, ``{\bf ID}'' represents the Wikipedia page's title of the sampled sentence. ``{\bf Sentence}'' indicates the sampled sentence. ``{\bf Triples to gold triples}'' column contains the triples (on the left side of ``$\rightarrow$'') and their corresponding gold triples (on the right side of ``$\rightarrow$'').

\begin{figure*}
\begin{center}
\tiny
\resizebox{0.85\linewidth}{!}
  {
\begin{tabular}{l p{11cm} p{11cm}}
\toprule
{\bf ID} & {\bf Sentence} & {\bf Triples to gold triples} \\ \hline
\input{appendices_table/mapped_bertlarge_kbp}
\bottomrule
\end{tabular}
}
\end{center}
\vspace{-0.1in}
\caption{{\small BERT$_{\rm LARGE}$ on TAC KBP-OIE.}} \label{fig:kbpmapbert}
\end{figure*}

\begin{figure*}
\begin{center}
\tiny
\resizebox{0.81\linewidth}{!}
  {
\begin{tabular}{l p{11cm} p{11cm}}
\toprule
{\bf ID} & {\bf Sentence} & {\bf Triples to gold triples} \\ \hline
\input{appendices_table/mapped_gpt2xl_kbp}
\bottomrule
\end{tabular}
}
\end{center}
\caption{{\small GPT-2$_{\rm XL}$ on TAC KBP-OIE.}} \label{fig:kbpmapgpt}
\end{figure*}

\begin{figure*}
\begin{center}
\tiny
\resizebox{0.85\linewidth}{!}
  {
\begin{tabular}{l p{7cm} p{7cm} p{7cm}}
\toprule
\multirow{2}{*}{\bf ID} & \multirow{2}{*}{\bf Sentence} & \multicolumn{2}{c}{\bf Triples to gold triples} \\
& & {\bf BERT$_{\rm LARGE}$} & {\bf GPT-2$_{\rm XL}$} \\ \hline
\input{appendices_table/mapped_compare_kbp}
\bottomrule
\end{tabular}
}
\end{center}
\caption{{\small BERT$_{\rm LARGE}$ vs. GPT-2$_{\rm XL}$ on TAC KBP-OIE.}} \label{fig:kbpmapbertgpt}
\end{figure*}

\begin{figure*}
\begin{center}
\tiny
\resizebox{0.81\linewidth}{!}
  {
\begin{tabular}{l p{10cm} p{10cm}}
\toprule
{\bf ID} & {\bf Sentence} & {\bf Triples to gold triples} \\ \hline
\input{appendices_table/mapped_bertlarge_wikidata}
\bottomrule
\end{tabular}
}
\end{center}
\caption{{\small BERT$_{\rm LARGE}$ on Wikidata-OIE.}} \label{fig:wikimapbert}
\end{figure*}

\begin{figure*}
\begin{center}
\tiny
\resizebox{0.85\linewidth}{!}
  {
\begin{tabular}{l p{10cm} p{10cm}}
\toprule
{\bf ID} & {\bf Sentence} & {\bf Triples to gold triples} \\ \hline
\input{appendices_table/mapped_gpt2xl_wikidata}
\bottomrule
\end{tabular}
}
\end{center}
\caption{{\small GPT-2$_{\rm XL}$ on Wikidata-OIE.}} \label{fig:wikimapgpt}
\end{figure*}

\begin{figure*}
\begin{center}
\tiny
\resizebox{0.85\linewidth}{!}
  {
\begin{tabular}{l p{7cm} p{7cm} p{7cm}}
\toprule
\multirow{2}{*}{\bf ID} & \multirow{2}{*}{\bf Sentence} & \multicolumn{2}{c}{\bf Triples to gold triples} \\
& & {\bf BERT$_{\rm LARGE}$} & {\bf GPT-2$_{\rm XL}$} \\ \hline
\input{appendices_table/mapped_compare_wikidata}
\bottomrule
\end{tabular}
}
\end{center}
\caption{{\small BERT$_{\rm LARGE}$ vs. GPT-2$_{\rm XL}$ on Wikidata-OIE.}} \label{fig:wikimapbertgpt}
\end{figure*}

\end{document}